\documentclass[twoside,11pt]{article}

\usepackage{blindtext}

\usepackage{jmlr2e}
\usepackage{subfigure}
\usepackage{bbding}
\usepackage{bm}
\usepackage{mathtools}
\usepackage{multirow}
\usepackage[ruled,linesnumbered]{algorithm2e}
\usepackage{enumitem}
\usepackage[normalem]{ulem}
\usepackage{float}
\usepackage{booktabs}

\newcommand{\specialcell}[2][c]{%
	\begin{tabular}[#1]{@{}c@{}}#2\end{tabular}}

\usepackage{lastpage}

\firstpageno{1}

\begin{document}

\title{Generalizing Hyperedge Expansion for Hyper-relational Knowledge Graph Modeling}

\author{\name Yu Liu \email liuyu2419@126.com\\
       \addr Department of Electronic Engineering\\
       Tsinghua University\\
       Beijing, China
              \AND
       \name Shu Yang \email shu-yang18@mails.tsinghua.edu.cn \\
       \addr Department of Electronic Engineering\\
       Tsinghua University\\
       Beijing, China
       \AND
		\name Jingtao Ding \email dingjt15@tsinghua.org.cn \\
		\addr Department of Electronic Engineering\\
		Tsinghua University\\
		Beijing, China
       \AND
       \name Quanming Yao \email qyaoaa@tsinghua.edu.cn \\
       \addr Department of Electronic Engineering\\
       Tsinghua University\\
       Beijing, China
       \AND
       \name Yong Li \email liyong07@tsinghua.edu.cn \\
       \addr Department of Electronic Engineering\\
       Tsinghua University\\
       Beijing, China
   }

\editor{My editor}

\maketitle

\begin{abstract}%
		By representing knowledge in a primary triple associated with additional attribute-value qualifiers, hyper-relational knowledge graph (HKG) that generalizes triple-based knowledge graph (KG) has been attracting research attention recently. Compared with KG, HKG is enriched with the semantic qualifiers as well as the hyper-relational graph structure. However, to model HKG, existing studies mainly focus on either semantic information or structural information therein, which however fail to capture both simultaneously. To tackle this issue, in this paper, we generalize the hyperedge expansion in hypergraph learning and propose an equivalent transformation for HKG modeling, referred to as TransEQ. Specifically, the equivalent transformation transforms a HKG to a KG, which considers both semantic and structural characteristics. Then an encoder-decoder framework is developed to bridge the modeling research between KG and HKG. In the encoder part, KG-based graph neural networks are leveraged for structural modeling; while in the decoder part, various HKG-based scoring functions are exploited for semantic modeling. Especially, we design the sharing embedding mechanism in the encoder-decoder framework with semantic relatedness captured. We further theoretically prove that TransEQ preserves complete information in the equivalent transformation, and also achieves full expressivity. Finally, extensive experiments on three benchmarks demonstrate the superior performance of TransEQ in terms of both effectiveness and efficiency. On the largest benchmark WikiPeople, TransEQ significantly improves the state-of-the-art models by 15\% on MRR. 
\end{abstract}


	\section{Introduction}\label{sec:intro}
	
	In the past decade, knowledge graph (KG) has been widely studied in artificial intelligence area \citep{wang2017knowledge,ji2020survey}. 
	By representing facts into a triple of $(s,r,o)$ with subject entity $s$, object entity $o$ and relation $r$, KG stores real-world knowledge in a graph structure. 
	However, recent studies find that KG with simple triples provides incomplete information \citep{StarE,yu2021improving,xiong2023reasoning}. 
	For example, both $($\emph{Alan Turing}, \texttt{educated at}, \emph{Cambridge}$)$ and $($\emph{Alan Turing}, \texttt{educated at}, \emph{Princeton}$)$ are true facts in KG, which might be ambiguous when the degree matters. 
	
\begin{figure}[htbp]
         \centering
		\includegraphics[width=0.6\linewidth]{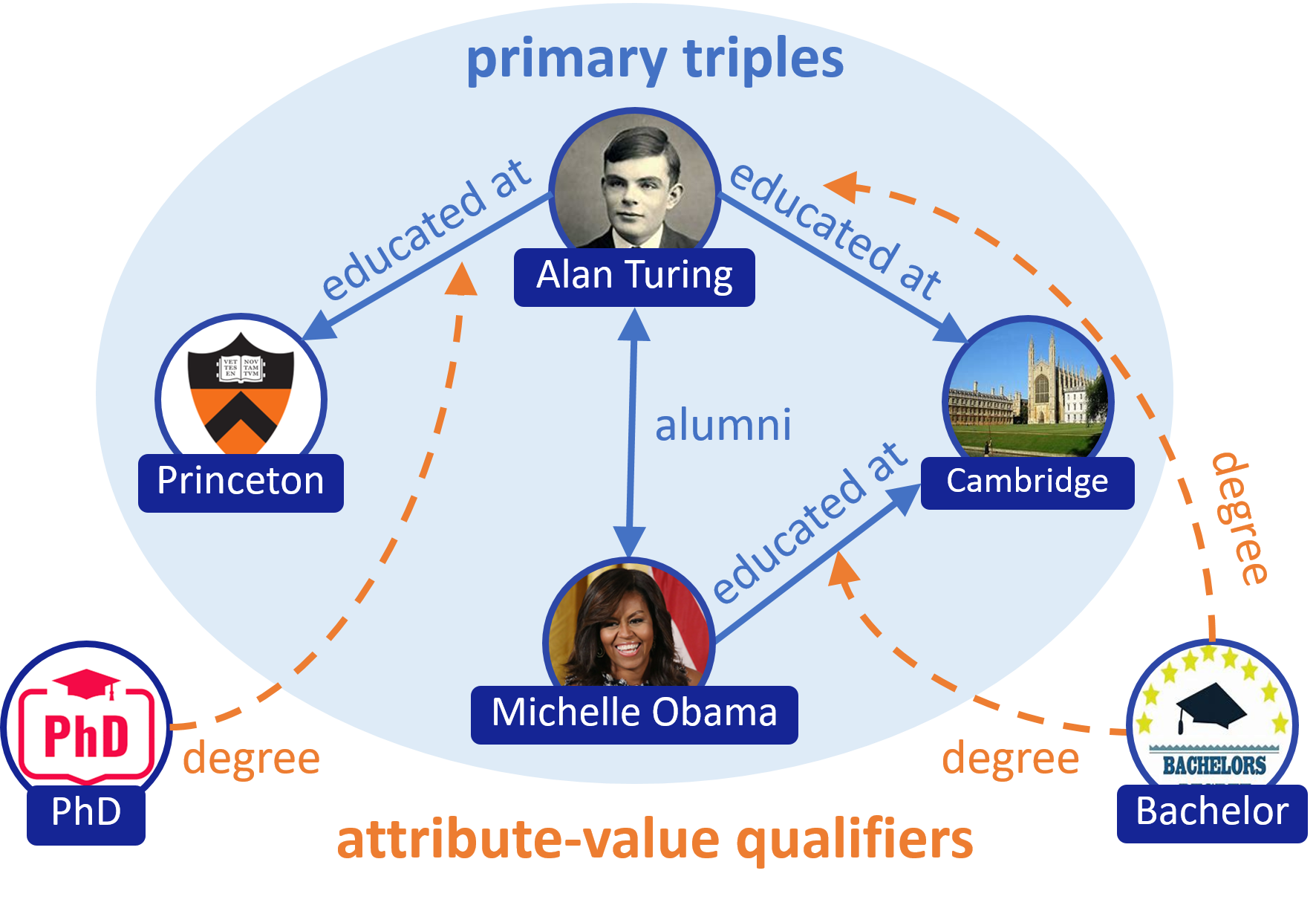}
		\vspace{-5px}
		\caption{An example of a HKG including primary triples and attribute-value qualifiers.}
		\label{fig:example}
	\end{figure}

	Hence, the hyper-relational KG (HKG) \citep{StarE}, a.k.a., knowledge hypergraph \citep{fatemi2019knowledge} and n-ary knowledge base \citep{guan2019link,liu2021role}, is proposed for more generalized knowledge representation. 
	Formally, in HKG, a primary triple is augmented with additional attribute-value qualifiers for rich semantics, called the hyper-relational fact\footnote{For a hyper-relational fact, entities/relation in the triple are called as primary entities/relation, and attributes/values in qualifiers are called as qualifier entities/relations. Note that the triple without qualifiers is a special case of hyper-relational facts.} \citep{guan2020neuinfer}. 
	Taking Figure~\ref{fig:example} as an example, both $($\emph{Alan Turing}, \texttt{educated at}, \emph{Cambridge}, $($\texttt{degree}, \emph{Bachelor}$)$$)$ and (\emph{Alan Turing}, \texttt{educated at}, \emph{Princeton}, $($\texttt{degree}, \emph{PhD}$)$$)$ are hyper-relational facts, where (\texttt{degree}, \emph{Bachelor}) and (\texttt{degree}, \emph{PhD}) are qualifiers with the degree attribute considered. 
	Such hyper-relational facts are ubiquitous that over 1/3 of the entities in Freebase \citep{bollacker2008freebase} involve in them \citep{wen2016representation}.
	
	To learn from HKG and further benefit the downstream tasks, HKG modeling learns low-dimensional vector representations (embeddings) of entities and relations \citep{guan2019link,chung2023representation}, which designs a scoring function (SF) based on the embeddings to measure the hyper-relational fact plausibility such that valid ones obtain higher scores than invalid ones. 
	Especially, existing studies mainly consider two aspects of semantic information and structural information in HKG for modeling. 
	
	The semantic information emphasizes the interaction between entities and relations in a hyper-relational fact. 
	Especially, there is a distinction, a.k.a., semantic difference \citep{StarE} between the primary triple and attribute-value qualifiers according to the occurrence frequency in world knowledge. For example, the primary triple $($\emph{Alan Turing}, \texttt{educated at}, \emph{Cambridge}$)$ serves as the fundamental part and preserves the essential knowledge of Alan Turing's education experience at Cambridge, while the attribute-value qualifier $($\texttt{degree}, \emph{Bachelor}$)$ serves as the auxiliary part and enriches the primary triple. 
	To model the semantic information, 
	early studies treat the primary relation and qualifier relations as an n-ary (n$\geq$2) composed relation \citep{fatemi2019knowledge,abboud2020boxe,wang2023hyconve} or multiple semantically equal attributes \citep{guan2019link,liu2021role}, largely ignoring the semantic difference. 
	Various SFs are further developed in recent studies \citep{StarE,guan2020neuinfer,rossobeyond} with such semantic difference considered. 
	
	On the other hand, the structural information focuses on the topological connection between entities in the hyper-relational graph structure\footnote{The HKG is also known as in a multi-relational hypergraph structure \citep{yadati2020neural}, and here we adopt the commonly used term ``hyper-relational graph structure'' in existing studies \citep{StarE}.}, like an entity's neighboring entities under various hyper-relational links, e.g., in Figure~\ref{fig:example} \emph{Bachelor} and \emph{Michelle Obama} are neighbors of \emph{Alan Turing} via \texttt{degree} and \texttt{alumni}, respectively. 
	Only few studies \citep{StarE,yadati2020neural} extend hypergraph neural network (HGNN) based modules to capture the structural information in HKG, however, empirical results in \citep{yu2021improving} demonstrate that removing such modules will not bring performance degradation, i.e., the direct extensions are quite immature for effective structural information capture. 
	Hence, to the best of our knowledge, none of existing studies achieve HKG modeling with both semantic information and structural information completely captured, and it is still an open problem to be addressed. 
	
	Targeting this open problem, we look back to KG modeling with an interesting observation that, recent studies \citep{schlichtkrull2018modeling,vashishth2019composition,yu2021knowledge} leverage an encoder-decoder framework for KG modeling, i.e., a powerful graph neural network (GNN) based encoder and an expressive SF-based\footnote{Here the SF refers to scoring function proposed in KG modeling with only triples considered, e.g., TransE \citep{bordes2013translating}, DistMult \citep{yang2014embedding}, etc.} decoder are leveraged for structural information and semantic information, respectively. On the other hand, in hypergraph learning, the hyperedge expansion is widely used to transform a hypergraph to a graph \citep{zhou2006learning,antelmi2023survey}. 
	Inspired by these two points, in this paper, we propose an \underline{EQ}uivalent \underline{Trans}formation for HKG modeling, termed as TransEQ. Specifically, TransEQ firstly generalizes the hyperedge expansion to an equivalent transformation, transforming a HKG to a KG, based on which an encoder-decoder framework is further developed to capture information. 
	For structural information, TransEQ introduces a GNN-based encoder on transformed KG with transformation characteristics combined. 
	As for semantic information, to measure the plausibility of a hyper-relational fact, TransEQ exploits various SFs in existing HKG modeling studies as the decoder. The sharing embedding mechanism is further designed to capture the semantic relatedness between hyper-relational facts.
	In this way,  with a simple yet effective transformation, the encoder-decoder framework in TransEQ captures not only structural information but also semantic information, which is the very innovation of this work, just like killing two birds with one stone.
	Besides, the flexible choice of SF in decoder ensures the full expressivity of TransEQ, representing all types of relations. 
	We further theoretically prove that the generalized transformation is equivalent between a HKG and a KG without information loss. 
	
	Our contributions are summarized as follows: 
	\begin{itemize}%
		\item We propose the TransEQ model for HKG modeling, which generalizes the hyperedge expansion to an equivalent transformation with a HKG transformed to a KG, and then develops an encoder-decoder framework to associate KG modeling research with HKG ones, capturing both structural information and semantic information. 
		\item We theoretically prove that the proposed equivalent transformation brings no information loss, i.e., the conversion between HKG and KG is equivalent. 
		We further prove that TransEQ is fully expressive to represent any HKG, completely separating valid facts from invalid ones.   
		\item We conduct extensive experiments and show that TransEQ achieves the state-of-the-art results across benchmarks, obtaining a 15\% relative increase of MRR on the largest benchmark WikiPeople. 
		Several studies demonstrate the model effectiveness and efficiency, and visualization results further indicate that TransEQ successfully captures the semantics in HKG. 
	\end{itemize}

	\section{Related Work}\label{sec:related_work}
	\subsection{Knowledge Graph (KG) Modeling}\label{sec:kg_learning}
	Learning representations for entities and relations in KGs has been investigated thoroughly \citep{wang2017knowledge,ji2020survey}, 
	which designs various SFs to model the semantics in triple knowledge $(s,r,o)$. 
	Based on translational thought, TransE \citep{bordes2013translating}, TransH \citep{TransH} and RotatE \citep{sun2019rotate} measure the distance between subject and object entities in a relation-specific latent space. 
	Besides, ConvE \citep{ConvE} adopts the convolutional neural networks for SF design. 
	TuckER \citep{balazevic2019tucker} employs Tucker decomposition for SF design. 
	Furthermore, 
	several models combine the bilinear product with various types of embeddings \citep{trouillon2016complex,cao2021dual,yang2014embedding}. 
	For example,  
	ComplEx \citep{trouillon2016complex} and DulE \citep{cao2021dual} employ complex-valued embeddings and dual quaternion embeddings, respectively.  
	However, models above ignore the multi-relational graph structure of KGs.
	
	Until the emergence of message passing mechanism with GNN, 
	structural information capture becomes an important topic in KG modeling. 
	An encoder-decoder framework is developed in recent KG-based GNN studies, where GNNs encode structural information of KG and various SFs are combined for semantic information. 
	Specifically, 
	both R-GCN \citep{schlichtkrull2018modeling} treats the multi-relational KG as multiple single-relational graphs, 
	and applies relational graph convolutional network (GCN) for entity representations. 
	Moreover, 
	VR-GCN \citep{ye2019vectorized} combines the translational idea with GNN to learn both entity and relation representations. 
	CompGCN \citep{vashishth2019composition} and InGram \cite{lee2023ingram} develop three entity-relation composition operators to update entity representations in GCN, 
	and KE-GCN \citep{yu2021knowledge} further incorporates the composition with relation update. 
	Overall, GNN-based models achieve promising results in KG modeling, 
	which demonstrates the importance of capturing structural information. 
	
	\subsection{Hyper-relational Knowledge Graph (HKG) Modeling}\label{sec:hkg_learning}
	
	As described before, related studies mainly exploit two aspects of semantic information and structural information for HKG modeling, considering HKG-based SF design and hyper-relational graph structure, respectively.

	\textbf{Semantic Modeling Studies.} For the semantic information, 
	given a hyper-relational fact $($\emph{Alan Turing}, \texttt{educated at}, \emph{Cambridge}, $($\texttt{degree}, \emph{Bachelor}$)$$)$, some studies \citep{wen2016representation,zhang2018scalable,abboud2020boxe,fatemi2019knowledge,fatemi2021knowledge,liu2020generalizing,wang2023hyconve} treat all involved relations as an n-ary composed relation \texttt{educated at\_degree} (here n is 3) with the fact $($\texttt{educated at\_degree}, \emph{Alan Turing}, \emph{Cambridge}, \emph{Bachelor}$)$. 
	These studies are directly extended from KG modeling methods without multiple relational semantics considered. 
	Especially, both m-TransH \citep{wen2016representation} and RAE \citep{zhang2018scalable} extend the SF of TransH \citep{TransH} to the hyper-relational case, while BoxE \citep{abboud2020boxe} combines translational idea with box embeddings. 
	Moreover, GETD \citep{liu2020generalizing} and S2S \citep{di2021searching} are both generalized from TuckER \citep{balazevic2019tucker}, where GETD further introduces tensor ring decomposition while S2S applies neural architecture search techniques. 
	The bilinear product is also extended to multilinear product with symmetric embeddings in m-DistMult \citep{yang2014embedding}, convolutional filters in HypE \citep{fatemi2019knowledge}, and relational algebra operations in ReAlE \citep{fatemi2021knowledge}. 
	On the other hand, NaLP \citep{guan2019link} and RAM \citep{liu2021role} decompose all involved relations into semantically equal attributes, and treat the example fact into a collection of attribute-value qualifiers, $($\texttt{educated at\_head}, \emph{Alan Turing}, \texttt{educated at\_tail}, \emph{Cambridge}, \texttt{degree}, \emph{Bachelor}$)$. 
	Nevertheless, models above largely ignore the semantic difference in hyper-relational facts. To capture the semantic difference between the primary triple and attribute-value qualifiers, NeuInfer \citep{guan2020neuinfer} and HINGE \citep{rossobeyond} design two sub-modules for HKG modeling, i.e., one for triple modeling and the other one for qualifier modeling, where NeuInfer mainly adopts fully connected layers while HINGE resorts to convolutional neural networks. Besides, GRAN \citep{wang2021link}, Hy-Transformer \citep{yu2021improving} and HyNT \citep{chung2023representation} leverage transformer and embedding processing techniques for HKG modeling. 
	However, these neural network based models rely on tremendous parameters for expressivity and are prone to overfitting.

	\textbf{Structural Modeling Studies.} As for the structural information, G-MPNN \citep{yadati2020neural} ignores attribute information and treats HKG as a multi-relational ordered hypergraph with n-ary composed relations, and further proposes multi-relational HGNN for modeling. 
	The rough design makes G-MPNN less competitive in practice. 
	StarE \citep{StarE} firstly introduces GNN for HKG modeling with a relation-specific message passing mechanism developed. 
	However, StarE aggregates hyper-relational fact messages for a specific entity only when the entity involves with the primary triple, but ignores the ones when the entity is in attribute-value qualifiers, i.e., StarE only captures connections among primary triples \citep{yu2021improving}. 
	Thus, capturing structural information for HKG modeling is still immature and needs further investigation. 
	
	Overall, existing HKG modeling studies are affected by various limitations from semantics and structure, while our proposed TransEQ elegantly models both aspects with full expressivity achieved, which is a quite important property for learning capacity in both KG modeling \citep{balazevic2019tucker,sun2019rotate} and HKG modeling \citep{fatemi2019knowledge,liu2020generalizing,abboud2020boxe}. 
	Besides, the inductive link prediction and logical query for HKG are investigated in recent studies \citep{ali2021improving,alivanistos2021query,chen2022explainable,lee2023ingram}, which are beyond the scope of this paper.

	\section{Preliminaries}\label{sec:preliminaries}
	\subsection{Hypergraph \& Hyperedge Expansion} \label{sec:hyper_expansion}
	A hypergraph is a generalization of graph, where a hyperedge can join any number of nodes \citep{ouvrard2020hypergraphs,antelmi2023survey}. 
	Especially, hyperedge expansion \citep{agarwal2006higher,zhou2006learning,dong2020hnhn} is introduced to transform a hypergraph to a homogeneous graph, such that graph learning methods can work on hypergraphs \citep{yadati2019hypergcn,feng2019hypergraph}. 
	Since HKG is viewed as a multi-relational hypergraph \citep{yadati2020neural}, here we investigate the representative expansion strategy of star expansion for additional insights. 
	
	\begin{figure}[htbp]
		\centering
		\subfigure[a hyperedge]{ \label{fig:plain_edge} 
			\includegraphics[width=0.25\textwidth]{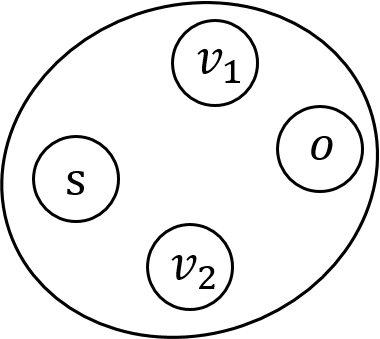}}
		\hspace{3em}
		\subfigure[star expansion]{ \label{fig:plain_star} 
			\includegraphics[width=0.25\textwidth]{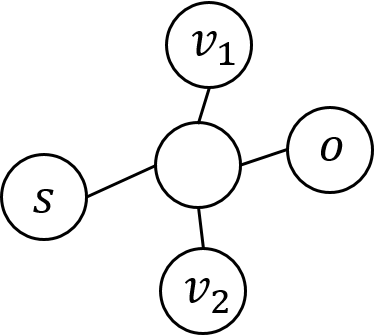}}
		\vspace{-5px}
		\caption{The illustration of star expansion on a hyperedge.}
		\label{fig:star_clique}
	\end{figure}
	
	Figure~\ref{fig:star_clique} presents an example of a hyperedge with four nodes and its transformed graph by star expansion. 
	Specifically, for a hyperedge, the star expansion introduces a mediator node (like the blank node in the center of Figure~\ref{fig:plain_star}), which is then connected with all original nodes in the hyperedge. 
	With the elegant transformation, hyperedge expansion has been widely applied in recommender systems \citep{xia2021self}, link prediction \citep{sun2021multi}, etc.

	On the other hand, the structural information loss has always been a concerned issue with hyperedge expansion strategy \citep{zhou2006learning,dong2020hnhn,antelmi2023survey}. 
	To be specific, an expansion strategy on hypergraph suffers from structural information loss, if there can be two distinct hypergraphs on the same node set reduced to the same graph by the expansion \citep{dong2020hnhn}. 
	According to \citep{dong2020hnhn}, the star expansion preserves the complete structural information. 
	However, existing hyperedge expansion strategies are focus on hypergraph, which cannot handle the HKG with hyper-relational semantics considered. Therefore, a generalization of hyperedge expansion to HKG is necessary to preserve both structural and semantic information therein.

	\subsection{Hyper-relational Knowledge Graph}
	Here we introduce the mathematical definition of HKG as well as the investigated problem. 
	\begin{definition}\label{def:hkg}
		\textbf{Hyper-relational Knowledge Graph.} 
		A HKG is defined as $\mathcal{G}^H=(\mathcal{E},\mathcal{R},\mathcal{F}^H)$, 
		where $\mathcal{E}$ and $\mathcal{R}$ are the sets of entities and relations, respectively. 
		A hyper-relational fact can be expressed as $(s,r,o,\{(a_i,v_i)\}^n_{i=1})$, where $(s,r,o)$ is the primary triple and $\{(a_i,v_i)\!\mid\! a_i\!\in\!\mathcal{R}, v_i\!\in\!\mathcal{E}\}^n_{i=1}$ is the attribute-value qualifier set.
		Moreover, $\mathcal{F}^H\!\subseteq\!\mathcal{E}\times\mathcal{R}\times\mathcal{E}\times\mathcal{P}$ denotes the fact set and $\mathcal{P}$ denotes all possible combinations of attribute-value qualifiers.
	\end{definition}
	
	Note that the number of qualifiers can be zero for a hyper-relational fact, i.e., HKG reduces to KG with an empty set $\mathcal{P}$. 
	In practice, attributes and values are also described by relations and entities, respectively \citep{StarE,yu2021improving}. 
	Then we state our research problem. 
	
	\newtheorem{problem}{Problem}
	\begin{problem}
		\textbf{\emph{HKG Modeling Problem.}} 
		Given a HKG $\mathcal{G}^H=(\mathcal{E},\mathcal{R},\mathcal{F}^H)$, the HKG modeling problem aims to learn representations for entities and relations in $\mathcal{E}$ and $\mathcal{R}$.
	\end{problem} 
	Especially, the HKG is always incomplete, which specifies the research problem as HKG completion problem in practice, i.e., given an incomplete hyper-relational fact with an entity missing at triple or qualifiers, inferring the missing entity from $\mathcal{E}$ with observable facts $\mathcal{F}^H$. 
	According to the definition, HKG involves semantic information of primary triple and attribute-value qualifiers as well as structural information of hyper-relational graph structure, which should be elegantly considered in modeling.

	\section{Method} \label{sec:method}

	As described before, the encoder-decoder framework has shown superior performance to capture both structural information and semantic information in KG \citep{schlichtkrull2018modeling,vashishth2019composition,lee2023ingram}, and thus a natural idea is to explore it for HKG modeling. Moreover, the hyperedge expansion studies provide a motivation to our work transforming a HKG to a KG with the encoder-decoder framework combined. Hence, we build TransEQ with such points in mind, which is presented in the following.

	\subsection{The TransEQ Model} \label{sec:TransEQ}
\begin{figure*}[htbp]
        \centering
		\includegraphics[width=0.95\linewidth]{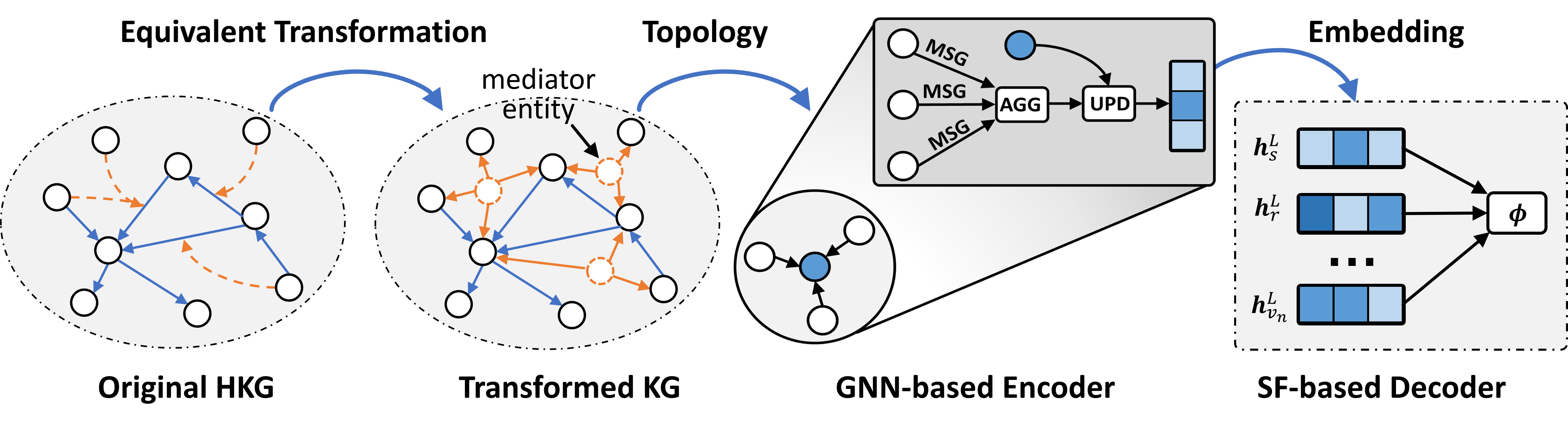}
		\vspace{-5px}
		\caption{The architecture of our proposed HKG modeling model TransEQ. An original HKG is transformed to a KG via the equivalent transformation, and then a GNN-based encoder and a SF-based decoder are leveraged for modeling structural information and semantic information, respectively.}
		\label{fig:sysmtem_framework}
	   \end{figure*}

	We now come to the details of our proposed TransEQ model, the architecture of which is illustrated in Figure~\ref{fig:sysmtem_framework}. 
	TransEQ first introduces the equivalent transformation with a HKG transformed to a KG, and then develops an encoder-decoder framework to capture both structural information and semantic information in HKG. 
	The model training process as well as other variants of transformations are also provided in this part.
	
	\subsubsection{One Stone: Equivalent Transformation} \label{sec:transformation}
	To identify the importance of transformation between HKG and KG, here we first introduce the definition of equivalent transformation. 
	
	\begin{definition}\label{def:equiv_trans}
		\textbf{Equivalent Transformation.} 
		A transformation between HKG and KG is equivalent, if the transformation preserves the complete information, i.e., given any HKG and its transformed KG via the transformation, they can be retrieved from each other. 
	\end{definition}

\begin{figure}[htbp]
        \centering
		\includegraphics[width=0.7\linewidth]{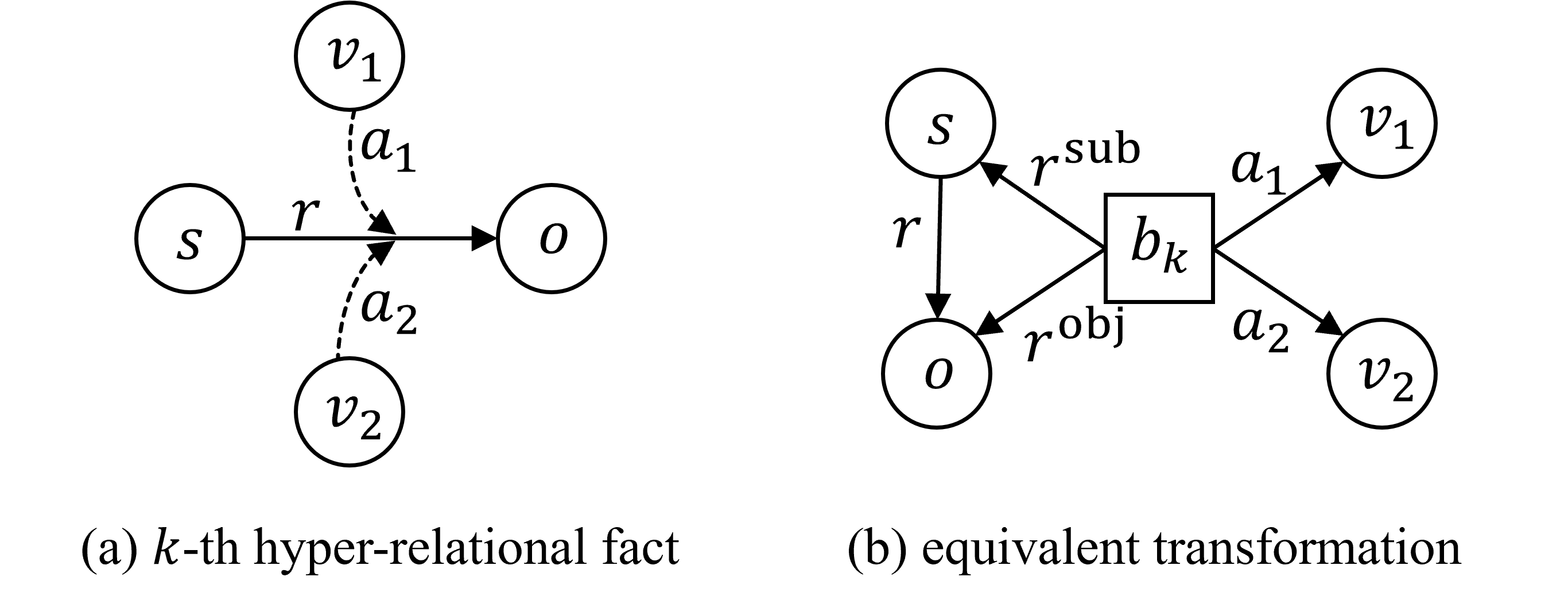}
		\vspace{-5px}
		\caption{The illustration of the equivalent transformation.}
		\label{fig:equiv_transformation}
	\end{figure}

    Moreover, a hyper-relational fact $(s,r,o,\{(a_i,v_i)\}^n_{i=1})$ can be viewed as a hyper-relational edge, which connects entities of $s,o,\{v_i\}^n_{i=1}$ with heterogeneous semantics of primary relation $r$ and attributes $\{a_i\}^n_{i=1}$, as shown in Figure~\ref{fig:equiv_transformation}(a) with the $k$-th fact in a HKG. 
	Thus, by generalizing star expansion as well as standard RDF reification in semantic web \citep{bollacker2008freebase,hernandez2015reifying,frey2019evaluation}, we propose an equivalent transformation for hyper-relational edges such that entities and relations in the original HKG are reorganized to the transformed KG with both structural information and semantic information preserved. 
 
	Specifically, the equivalent transformation in Figure~\ref{fig:equiv_transformation}(b) introduces a mediator entity $b_k$ to identify the fact, and the primary relation $r$ is extended with two relations $r^\text{sub}$ and $r^\text{obj}$ for the relational edges between $b_k$ and subject entity $s$ and object entity $o$, respectively. 
	The attribute information in original hyper-relational fact is preserved by the attributed-based edges between $b_k$ and value entities. 
	Moreover, a relational edge $r$ connects entities $s$ and $o$ for semantic difference, i.e., such operation leads to a three-node clique motif \citep{milo2002network}, 
	reflecting the primary role of the triple. Hence, compared with the star expansion in Figure~\ref{fig:plain_star}, the equivalent transformation incorporates the semantic information.
	
	For better understanding, we present the execution process of the equivalent transformation in Algorithm~\ref{alg:equiv_transformation}. 
	Especially, in lines 7-8, TransEQ utilizes different transformation operations to model the semantic difference. 
	Besides, the original structure of the triple fact, i.e., hyper-relational fact without qualifiers, is kept to avoid redundancy. 
	As proved later, such transformation brings no information loss, and provides a good basis for the following encoder-decoder framework. 
	
	\SetKwInput{KwInit}{Init}
	\begin{algorithm}[htbp]
		\caption{The algorithm for equivalent transformation.}\label{alg:equiv_transformation}
		
		\KwIn{HKG $\mathcal{G}^H=(\mathcal{E},\mathcal{R},\mathcal{F}^H)$; }
		
		\KwInit{Transformed KG $\mathcal{G}=(\mathcal{E},\mathcal{R},\mathcal{F})$ with $\mathcal{F}\leftarrow\emptyset$;}
		
		Obtain the set of primary relations for facts in $\mathcal{F}^H$, $\mathcal{R}^\textnormal{pri}$;\\
		
		\For{$r\in\mathcal{R}^\textnormal{pri}$}
		{Define new relations $r_s,r_o$, and $\mathcal{R}\leftarrow\mathcal{R}\cup\{r^\text{sub},r^\text{obj}\}$;}
		
		\For{$k$\textnormal{-th fact} $(s,r,o,\{(a_i,v_i)\}_{i=1}^n)\in\mathcal{F}^H$}{
			Define mediator entity $b_k$, and $\mathcal{E}\leftarrow\mathcal{E}\cup \{b_k\}$;\\
			
			$\mathcal{F} \leftarrow \mathcal{F} \cup \{(s,r,o),(b_k,r^\text{sub},s),(b_k,r^\text{obj},o)\}$;\\
			$\mathcal{F} \leftarrow \mathcal{F} \cup \{(b_k,a_i,v_i)\}^n_{i=1}$;\\
			\tcp{semantic difference modeling in lines 7-8.}	
		}
		\KwOut{Transformed KG $\mathcal{G}=(\mathcal{E},\mathcal{R},\mathcal{F})$.}
	\end{algorithm}
	
	\subsubsection{Two Birds: Encoder-Decoder Framework} \label{sec:encoder-decoder} 
	To model both structural information and semantic information in the original HKG, TransEQ further introduces an encoder-decoder framework on the transformed KG. 
	
	As for the encoder part, powerful GNN is developed to capture structural information, where the semantic relatedness in HKG and the mediator entities in equivalent transformation are also incorporated therein. 
	Especially, the semantic relatedness explicitly lies in the shared primary relations across hyper-relational facts. 
	For example, the hyper-relational facts of $($\emph{Alan Turing}, \texttt{educated at}, \emph{Cambridge}, $($\texttt{degree}, \emph{Bachelor}$)$$)$ and $($\emph{Alan Turing}, \texttt{educated at}, \emph{Princeton}, $($\texttt{degree}, \emph{PhD}$)$$)$ share the same primary relation, which indicates a strong semantic relatedness. 
	On the other hand, in our proposed equivalent transformation, each mediator entity plays an important role of relaying connections among the entities in an original hyper-relational fact, and thus mediator entities aggregate the semantics of corresponding facts. 
	
	Hence, we introduce the sharing embedding for mediator entities to capture the semantic relatedness, and further combine it with three steps of unified multi-relational message passing mechanism in KG-based GNN \citep{schlichtkrull2018modeling,vashishth2019composition}. 
	\begin{itemize}[leftmargin=*]
		\item \textit{Initialized embedding.}
		Given the embedding dimension $d$, for the mediator entity $b$, we denote $\psi(b)$ the mapping from $b$ to its involved primary relation, and initialize its representation as $\bm{h}_{b}^{0}\!=\![\bm{e}_{\psi(b)};\bm{e}_b]$, where $\bm{e}_{\psi(b)}\!\in\!\mathbb{R}^{\lfloor\alpha\cdot d\rfloor}$ and $\bm{e}_b\!\in\!\mathbb{R}^{d-\lfloor \alpha\cdot d\rfloor}$ are sharing and independent embeddings, respectively. $\alpha$ is the hyperparameter to tune the sharing embedding ratio. 
		Thus, mediator entities involved with the same primary relation $\psi(b)$ share part of embedding $\bm{e}_{\psi(b)}$. 
		\item \textit{Message calculation.}
		Considering the stacking layers of GNN, we denote $\bm{m}_{urt}^{l+1,\text{ent}}$ and $\bm{m}_{urt}^{l+1,\text{rel}}$ the messages from a triple $(u,r,t)$ for target entity $t$ and relation $r$ at the $(l+1)$-th layer, respectively, which are calculated as follows,
		\begin{align}
			\bm{m}^{l+1,\text{ent}}_{urt}=\text{MSG}^{\text{ent}}(\bm{h}^l_u,\bm{h}^l_r,\bm{h}^l_t),\
			\bm{m}^{l+1,\text{rel}}_{urt}=\text{MSG}^{\text{rel}}(\bm{h}^l_u,\bm{h}^l_r,\bm{h}^l_t),\notag
		\end{align}
		where $\bm{h}^l_{u},\bm{h}^l_{t},\bm{h}^l_{r}\!\in\!\mathbb{R}^d$ are the embeddings of entities and relation at the $l$-th layer, while $\text{MSG}^{\text{ent}}$ and $\text{MSG}^{\text{rel}}$ can be composition function in CompGCN \citep{vashishth2019composition}, relation-specific projection in R-GCN \citep{schlichtkrull2018modeling} and etc. 
		Besides, the entity representations at the input layer are expressed as, 
		\begin{align}
			\!\!\!
			\bm{h}^0_{x}=
			\begin{cases}
				[\bm{e}_{\psi(x)};\bm{e}_x] & \mbox{if  $x$ is mediator entity}\\
				\hspace{0.2em}	\bm{e}^\prime_x\in\mathbb{R}^d  &  \mbox{if  $x$ is original entity}
			\end{cases},
			\ \text{for} \ x\in\{u,t\},\notag
		\end{align}
		
		\item \textit{Message aggregation.}
		Then neighborhood messages of $\bm{M}^{l+1}_{t}$ and $\bm{M}^{l+1}_{r}$ are aggregated as follows, 
		\begin{align}
			& \bm{M}^{l+1}_{t}\!=\!\text{AGG}^\text{ent}(\bm{m}^{l+1,\text{ent}}_{urt}\!\mid\!r\!\in\!\mathcal{R}, u\!\in\!\mathcal{N}^r_t), \notag \\ & \bm{M}^{l+1}_{r}\!=\!\text{AGG}^\text{rel}(\bm{m}^{l+1,\text{rel}}_{urt}\!\mid\!(u,t)\!\in\!\mathcal{N}_r),\notag
		\end{align}
		where $\mathcal{N}^r_{t}$ denotes the entities linked to $t$ via relation $r$ and $\mathcal{N}_r$ denotes the entity pair linked by relation $r$. 
		$\text{AGG}^\text{ent}$ and $\text{AGG}^\text{rel}$ are aggregation functions such as mean and sum pooling functions. 
		
		\item \textit{Representation update.}
		Finally, the representations at the $(l+1)$-th layer are updated with aggregated messages and former layer representations:
		\begin{align}
			\bm{h}^{l+1}_t=\text{UPD}^{\text{ent}}(\bm{M}^{l+1}_t,\bm{h}^{l}_t),\
			\bm{h}^{l+1}_r=\text{UPD}^{\text{rel}}(\bm{M}^{l+1}_r,\bm{h}^{l}_r),\notag
		\end{align}
		where $\text{UPD}^\text{ent}$ and $\text{UPD}^\text{rel}$ can be nonlinear activation functions. 
	\end{itemize} 
	
	Owing to above encoding process, TransEQ fully exploits the topological connections between entities for structural information. 
	
	Furthermore, the decoder part exploits various SFs to model semantic information. 
	For each hyper-relational fact, the encoder part feeds the representations of corresponding entities and relations into the SF-based decoder to model the interaction between entities and relations therein. 
	Especially, the choice of SF is orthogonal to the encoder \citep{schlichtkrull2018modeling,vashishth2019composition}, and most existing SFs on HKG modeling can be modified in the decoder. 
	For example of the hyper-relational fact $x\!\coloneqq\!(s,r,o,\{(a_i,v_i)\}^n_{i=1})\!\in\!\mathcal{F}^H$, we rewrite m-DistMult's SF \citep{fatemi2019knowledge} as:
	\begin{align}
		\phi(x)=\langle\varphi(\bm{h}^L_r,\bm{h}^L_{a_1},\cdots,\bm{h}^L_{a_n}),\bm{h}^L_s,\bm{h}^L_o,\bm{h}^L_{v_1},\cdots,\bm{h}^L_{v_n} \rangle,\notag
	\end{align} 
	where $\phi(x)$ is plausibility score measured by TransEQ, $\left\langle\cdot\right\rangle$ denotes the multilinear product\footnote{$\left\langle\bm{h}_{1},\bm{h}_{2},\cdots,\bm{h}_{n}\right\rangle=\sum_i \bm{h}_{1}[i]$ $\bm{h}_{2}[i]  \cdots\bm{h}_{n}[i]$}, and $L$ denotes the number of GNN layers in encoder part. 
	Since m-DistMult adopts a composed relation for SF, we introduce the function $\varphi$ to aggregate the embeddings of involved primary relation and attributes for the composed relation embedding, such as mean/sum pooling function. 
	Note that the semantic difference in HKG is also modeled by the SF in decoder. 
	Various SFs are further investigated in experiments later.

	\subsubsection{Model Training}\label{sec:training}

 	\begin{algorithm}[htbp]
		\caption{TransEQ training algorithm.} \label{alg:training}
		\KwIn{HKG $\mathcal{G}^H=(\mathcal{E},\mathcal{R},\mathcal{F}^H)$;}
		
		\KwInit{$\bm{E}$ for $e\in\mathcal{E}$, $\bm{R}$ for $r\in\mathcal{R}$, $\bm{\theta}_{\textnormal{Enc}}$ for GNN-based encoder, $\bm{\theta}_{\textnormal{Dec}}$ for SF-based decoder;}
		
		Build encoder module $Enc()$ with $\bm{\theta}_{\textnormal{Enc}}$;\\
		Build decoder module $Dec()$ with $\bm{\theta}_{\textnormal{Dec}}$;\\
		
		Transform HKG $\mathcal{G}^H$ to KG $\mathcal{G}$ with Algorithm~\ref{alg:equiv_transformation};\\
		
		\For{$t=1,\cdots,n_\textnormal{iter}$}{
			
			Sample a mini-batch $\mathcal{F}_{\text{batch}}\in \mathcal{F}^H$ of size $m_b$, $\mathcal{L}\!\leftarrow\! 0$; 		\\
			
			$\bm{E},\bm{R}=Enc(\mathcal{G},\bm{E},\bm{R},\bm{\theta}_\textnormal{Enc})$;\\
			\For{ $x\coloneqq(s,r,o,\{(a_i,v_i)\}^n_{i=1})\in\mathcal{F}_{\textnormal{batch}}$}
			{
				\raggedright{Construct negative samples $\mathcal{N}_x$;}
				\\
				
				\raggedright{$\phi(x)=Dec(x,\bm{E},\bm{R},\bm{\theta}_\textnormal{Dec})$;} 
				\\
				
				\raggedright{$\phi(x^\prime)=Dec(x^\prime,\bm{E},\bm{R},\bm{\theta}_\textnormal{Dec}), \ \forall x^\prime\in\mathcal{N}_x$;} 
				\\
				
				\raggedright{Update loss $\mathcal{L}\!\leftarrow\!\mathcal{L}\!+\!\mathcal{L}_x(\phi)$
					with $\mathcal{L}_x$ in \eqref{eq:loss};} 
				
			}
			\raggedright{Update learnable parameters w.r.t. the gradients $\nabla \mathcal{L}$;}
			
		}
		
		\KwOut{Embeddings $\bm{E},\bm{R}$ and parameters $\bm{\theta}_{\textnormal{Enc}},\bm{\theta}_{\textnormal{Dec}}$.}
	\end{algorithm}

	To learn the model parameters, we adopt the cross-entropy loss for training \citep{fatemi2019knowledge,yadati2020neural,liu2021role}. 
	For the hyper-relational fact $x\in\mathcal{F}^H$ with $\phi(x)$, the practical loss can be written as: 
	\begin{align}
		\mathcal{L} =\sum_{x\in\mathcal{F}^H}\!\!\mathcal{L}_x(\phi) = \sum_{x\in\mathcal{F}^H}\!\! -\log
		\frac{ e^{\phi(x)} }{ e^{\phi(x)}
			+
			\sum_{x^\prime\in\mathcal{N}_{x}} e^{\phi(x^\prime)}  }, 
		\label{eq:loss}
	\end{align}
	where $\mathcal{N}_x$ denotes the negative samples, i.e., entities in triple and attribute-value qualifiers of $x$ are replaced by other entities in $\mathcal{E}$. 
	Algorithm~\ref{alg:training} presents the training procedure of TransEQ. 
	The overall model is trained in a mini-batch way with batch normalization and dropout utilized for regularization.

	Overall, the proposed TransEQ develops an equivalent transformation that transforms a HKG to a KG. 
	Then an encoder-decoder framework associates KG modeling research with HKG ones, 
	where KG-based GNN encodes structural information while HKG-based SF in decoder focuses on semantic information. 
	
	\subsubsection{Other Variants of Generalized Transformations}\label{sec:other_transformation}
\begin{figure}[htbp]
    \centering
		\includegraphics[width=0.8\linewidth]{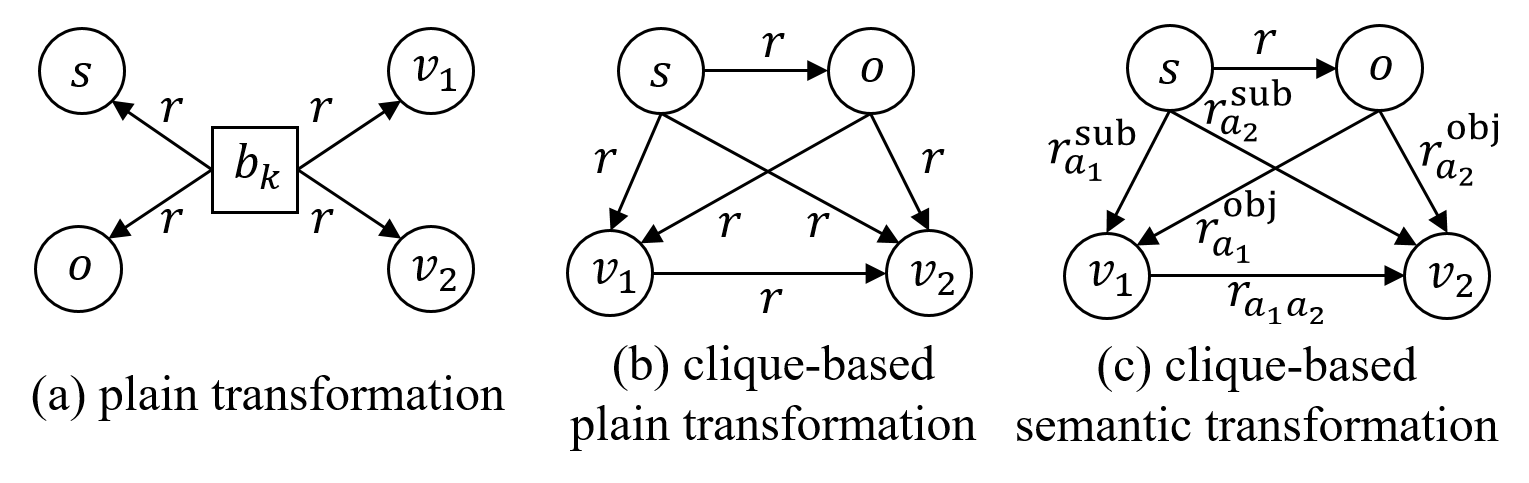}
		\vspace{-5px}
		\caption{The illustration of other variants of transformations.}
		\label{fig:other_transformation}
	\end{figure}

	To demonstrate the effectiveness of our generalized equivalent transformation in Section \ref{sec:transformation}, here we further show other variants of transformations in Figure~\ref{fig:other_transformation}. Especially, the plain transformation in Figure~\ref{fig:other_transformation}(a) follows star expansion without attributes considered. 
	
	\subsection{Theoretical Understanding}\label{sec:theoretical}

	\subsubsection{Complexity Analysis}\label{sec:complexity}
	To distinguish our proposed TransEQ model design, in Table~\ref{tab:models}, we present a comparison of HKG modeling studies with structural modeling, semantic modeling, full expressivity as well as time and space complexity.

	\begin{table}[htbp]
        \centering
		\caption{
			A comparison of representative HKG modeling studies. 
			$n_e$, $n_r$ and $n^\textmd{pri}_r$ denote the numbers of entities, relations and primary relations. 
			$d$ is the embedding dimension. 
			$n_a$ is the maximum number of attribute-value qualifiers for facts, and $N=\left|\mathcal{F}^H\right|$ is the total number of facts in HKG. HGNN: hypergraph neural network, GNN: graph neural network. 
			Neural: neural network based SF, Multilinear: multilinear product based SF.
		}
		\label{tab:models}
		\setlength\tabcolsep{1pt}
		\def\arraystretch{0.9}
		\begin{tabular}{cccccccc}
			\toprule
			\textbf{Model} & \specialcell{\textbf{Structure}\\ \textbf{Modeling}} & \specialcell{\textbf{Semantic}\\ \textbf{Difference}} & \specialcell{\textbf{Scoring}\\ \textbf{Function}}   &  \specialcell{\textbf{Expre}\\ \textbf{-ssive}} & $\bm{\mathcal{O}}_\text{time}$  & $\bm{\mathcal{O}}_\text{space}$ \\ 
			\midrule
			NaLP & \XSolidBrush & \XSolidBrush  & Neural & \XSolidBrush & $\mathcal{O}(d^2)$ & $\mathcal{O}(n_ed+n_rd)$ \\
			m-DistMult  & \XSolidBrush & \XSolidBrush   & Multilinear & \XSolidBrush  & $\mathcal{O}(d)$ & $\mathcal{O}(n_ed+n^{\text{pri}}_rd)$ \\
			HypE  & \XSolidBrush & \XSolidBrush   & Multilinear & \Checkmark & $\mathcal{O}(d)$ & $\mathcal{O}(n_ed+n^\text{pri}_rd)$\\
			HINGE   & \XSolidBrush & \Checkmark   & Neural &   \XSolidBrush & $\mathcal{O}(d^2)$ & $\mathcal{O}(n_ed+n_rd)$ \\
			G-MPNN  & HGNN & \XSolidBrush   & Multilinear & \XSolidBrush & $\mathcal{O}(Nd^2)$ & $\mathcal{O}(n_ed+n_r^\text{pri}d+n_ad)$ \\
			StarE  &  GNN  &\Checkmark   & Neural & \XSolidBrush & $\mathcal{O}(Nd^2+n_ad^2)$ & $\mathcal{O}(n_ed+n_rd)$ \\
			\midrule
			TransEQ & \specialcell{Transformation\\ \& GNN} &\Checkmark	&
			Arbitrary SF & \Checkmark & $\mathcal{O}(Nd^2)$ & $\mathcal{O}(n_ed+n_rd+Nd)$  \\ 
			\bottomrule
		\end{tabular}
	\end{table}

	According to the table, structural information is rarely explored in existing studies, while HGNN in G-MPNN is at its early stage and thus fails to model attribute semantics. 
	StarE only captures triple-based connections \citep{yu2021improving}, while TransEQ combines the equivalent transformation with GNN for structural information. 
	As for modeling semantic information, existing studies follow three views of relations in HKG, and adopt various SFs to model the interaction between entities and relations. 
	Especially, TransEQ can utilize arbitrary SF with semantic difference considered.
	Compared with the weak expressive power of most studies, the flexible choice of SF guarantees the full expressivity of TransEQ to model various HKGs, and brings performance improvement.  
	
	Besides, the message passing mechanism in modeling structural information leads to the time complexity of $\mathcal{O}(Nd^2)$, and Transformer module in StarE brings an additional complexity of $\mathcal{O}(n_ad^2)$. 
	Since the equivalent transformation introduces a mediator entity for each hyper-relational fact, TransEQ builds the space complexity of $\mathcal{O}(n_ed+n_rd+Nd)$ with original entities and relations considered. 
	Owing to GPU implementation, TransEQ obtains comparable efficiency to the fastest current studies in experiments.  
	In this way, TransEQ achieves efficient and expressive HKG modeling with both structural information and semantic information captured.

	\subsubsection{Information Preserving Transformation}\label{sec:info_loss}
	Following the structural information loss concern in hyperedge expansion \citep{zhou2006learning,dong2020hnhn,arya2021adaptive}, here we investigate the information loss problem for our generalized transformation on HKG, which emphasizes on preserving both structural information and semantic information. 
	
	Based on the equivalent transformation in Definition~\ref{def:equiv_trans} and the generalized transformation in TransEQ, we identify the property with the following theorem, and provide the proof in Appendix~\ref{app:info_loss}.  
    \setcounter{theorem}{0} 
	\begin{theorem}\label{theorem:loss}
		In the conversion from a HKG to a KG, the generalized transformation in TransEQ is an equivalent transformation and preserves the complete information, while other variants of transformations like plain transformation and clique-based transformations can lead to partial information loss.
	\end{theorem}

	\subsubsection{Full Expressivity}
	To demonstrate the expressivity of TransEQ, here we introduce the full expressivity property \citep{fatemi2019knowledge,liu2020generalizing,abboud2020boxe,fatemi2021knowledge,di2021searching}. 
	A HKG modeling model is fully expressive if, for any given HKG, the model can separate valid hyper-relational facts from invalid ones by appropriate parameter configuration. 
	
	Considering the encoder-decoder framework in TransEQ, such property is mainly determined by the SF in decoder part, thus we establish the expressivity of TransEQ with the following theorem.
	\begin{theorem}\label{theorem:expressive}
		With encoder parameters configured appropriately, the expressivity of TransEQ is in accord with that of the scoring function it uses in decoder, i.e., TransEQ is fully expressive if the scoring function used in decoder is fully expressive.
	\end{theorem}
	The proof is provided in Appendix~\ref{app:expressive}. 
	Thus, with appropriate choice of SF like HypE \citep{fatemi2019knowledge} as well as model parameters, a fully expressive TransEQ model has the potential to represent all types of relations in HKG including symmetric relations, inverse relations, etc. \citep{sun2019rotate,liu2021role}, which generally outperforms the weak ones in practice, as validated in Section~\ref{sec:encoder_decoder}.

	\section{Experiments and Results}\label{sec:experiment}
	\subsection{Experimental Setup}\label{sec:setup}
	\subsubsection{Datasets}
	The experiments are conducted on three benchmark HKG datasets, i.e., WikiPeople \citep{guan2019link}, JF17K \citep{zhang2018scalable} and FB-AUTO \citep{fatemi2019knowledge}. 
	We follow the original splits of WikiPeople and FB-AUTO, while JF17K follows the split in RAM \citep{liu2021role} with validation set further split. 
	The detailed statistics are summarized in Table~\ref{tab:datasets}. Details are provided in Appendix~\ref{app:exp_details}.	
	
	\begin{table}[htbp]
		\centering
        \caption{Dataset statistics in experiments.}
		\label{tab:datasets}
        \vspace{5px}
		\begin{tabular}{cccccc}
			\toprule
			Dataset & $\vert\mathcal{E}\vert$ & $\vert\mathcal{R}\vert$  & \#Train & \#Valid & \#Test \\ \midrule
			WikiPeople   & 47,765 & 707   & 305,725 & 38,223  & 38,281  \\
			JF17K  &  28,645  & 322  &  61,104  & 15,275  & 24,568 \\
			FB-AUTO & 3,388 & 8 &  6,778  &  2,255  & 2,180 \\ \bottomrule
		\end{tabular}
	\end{table}

	\subsubsection{Baselines}
	As for performance comparison, we compare with several state-of-the-art HKG modeling approaches, 
	including semantic modeling ones of m-TransH \citep{wen2016representation}, HypE \citep{fatemi2019knowledge}, RAM \citep{liu2021role}, S2S \citep{di2021searching}, HINGE \citep{rossobeyond}, NeuInfer \citep{guan2020neuinfer}, BoxE \citep{abboud2020boxe}, HyConvE \citep{wang2023hyconve} as well as structural modeling ones of StarE \citep{StarE}, G-MPNN \citep{yadati2020neural}.

	\subsubsection{Task and Evaluation Metrics}\label{sec:task_metrics}
	Following typical settings \citep{liu2020generalizing,abboud2020boxe,fatemi2019knowledge,guan2019link,wang2021link}, 
	we evaluate HKG modeling approaches on HKG completion task in transductive setting, and predict the missing entity at each position. 
	i.e., for a testing sample $(s,r,o,\{(a_i,v_i)\}^n_{i=1})$, we assume that either entity in $\{s,o,v_1,\cdots,v_n\}$ may be missing, and apply the model to predict missing entity. Note that this task is more generalized than previous study StarE \citep{StarE} only predicting missing entity in $\{s, p\}$ of triple part. As for evaluation metrics, the standard mean reciprocal ranking (MRR) and Hit@\{1,3,10\} are utilized in filtered setting \citep{bordes2013translating}.

	\subsubsection{Implementation}
	We implement TransEQ in PyTorch \citep{paszke2019pytorch} with Adam optimizer. 
	The embedding dimension $d$ is set to the typical size 200 \citep{wang2021link,abboud2020boxe,fatemi2019knowledge,yu2021improving,StarE}. 
	The batch size, learning rate and dropout are chosen from \{64, 128\}, \{0.0001, 0.0005, 0.001, 0.005\} and [0.1, 0.5] with step 0.1, respectively. 
	Besides, we mainly adopt CompGCN \citep{vashishth2019composition} as encoder and m-DistMult \citep{fatemi2019knowledge} as decoder. 
	For the encoder part, the number of GNN layers and sharing ratio $\alpha$ are chosen from \{1, 2, 3, 4\} and [0.0, 1.0] with step 0.2, respectively. 
	The composition operation in encoder is set to rotate function \citep{sun2019rotate}. 
	We tune hyperparameters over the validation set with early stopping strategy employed. 
	All experiments are run on a RTX 2080 Ti GPU.

	\begin{table}[htbp]
		\caption{Results of HKG completion on all datasets. 
			Results of baselines are collected from original papers and \citep{fatemi2019knowledge,liu2021role,anonymous2022message}. 
			Best results are highlighted in bold, and second best results are highlighted with underlines. "-" denotes missing results.}
		\label{tab:main_results}
		\def\arraystretch{0.9}
        \setlength\tabcolsep{1pt}
		\begin{tabular}{c|cccc|cccc|cccc}
			\toprule
			\multicolumn{1}{l}{} &\multicolumn{4}{c}{\textbf{WikiPeople}}&\multicolumn{4}{c}{\textbf{JF17K}} & \multicolumn{4}{c}{\textbf{FB-AUTO}}\\
			Model & MRR  & Hit@1  & Hit@3 & \multicolumn{1}{c|}{Hit@10} & MRR & Hit@1  & Hit@3 & \multicolumn{1}{c|}{Hit@10} & MRR  & Hit@1  & Hit@3  & Hit@10  \\ 
			\midrule
			m-TransH   & -  & -  &  -  & \multicolumn{1}{c|}{}   & 0.444   & 0.370  &  0.475   & \multicolumn{1}{c|}{0.581}  & 0.728  & 0.727  & 0.728  & 0.728 \\
			HINGE & 0.333 & 0.259  & 0.361 & \multicolumn{1}{c|}{0.477} & 0.473 & 0.397 & 0.490 & \multicolumn{1}{c|}{0.618} & 0.678 & 0.630  & 0.706  & 0.765  \\
			NeuInfer & 0.350  & 0.282   & 0.381 & \multicolumn{1}{c|}{0.467} & 0.517 & 0.436  & 0.553  & \multicolumn{1}{c|}{0.675} & 0.737  & 0.700 & 0.755   & 0.805  \\
			m-DistMult  & 0.318   & 0.213   & 0.391 & \multicolumn{1}{c|}{0.478}   & 0.452  & 0.375                & 0.482 & \multicolumn{1}{c|}{0.599}  & 0.784  & 0.745  & 0.815  & 0.845   \\
			HypE  & 0.292   & 0.162& 0.375  & \multicolumn{1}{c|}{0.502} & 0.507 & 0.421 & 0.550 & \multicolumn{1}{c|}{0.669} & 0.804 & 0.774  & 0.823 & 0.856  \\
			RAM  & 0.370 & \underline{0.293} & 0.410  & \multicolumn{1}{c|}{0.507} & 0.539 & 0.463  & 0.573  & \multicolumn{1}{c|}{\underline{0.690}} & 0.830  & 0.803 & 0.851  & 0.876   \\
			S2S & 0.372  & 0.277  & 0.439  & \multicolumn{1}{c|}{0.533} & 0.528  & 0.457  & 0.570  & \multicolumn{1}{c|}{\underline{0.690}} & - & - & -  &  -  \\
			BoxE & \underline{0.395}   & \underline{0.293} & \underline{0.453}  & \multicolumn{1}{c|}{0.503}  & \underline{0.560} & \underline{0.472} & \textbf{0.606} & \multicolumn{1}{c|}{\textbf{0.722}} & {0.844} & {0.814} & {0.863} & {0.898}  \\
            HyConvE & 0.362 & 0.275 & 0.388 & 0.501 & - & - & - & - & \underline{0.847} & \underline{0.820} & \underline{0.872} & \underline{0.901} \\
   
			\midrule
			G-MPNN  & 0.367 & 0.258 & 0.439 &  \multicolumn{1}{c|}{0.526}  & 0.530  & 0.459  & 0.572 & \multicolumn{1}{c|}{0.688} & 0.763 &  0.724  & 0.778  & 0.838  \\
			StarE & 0.378 & 0.265 & 0.452 &  \multicolumn{1}{c|}{\underline{0.542}} & {0.542} & 0.454 & 0.580 &  \multicolumn{1}{c|}{0.685} & 0.764 & 0.725 & 0.779 & 0.838\\
			\midrule\midrule
			TransEQ  &\textbf{0.454} & \textbf{0.373} & \textbf{0.495} & \multicolumn{1}{c|}{\textbf{0.593}} & \textbf{0.569} & \textbf{0.489} & \underline{0.601} & \multicolumn{1}{c|}{\textbf{0.722}} & \textbf{0.870} & \textbf{0.842} & \textbf{0.879} & \textbf{0.909} \\ 
			\bottomrule
		\end{tabular}
	\end{table}

	\subsection{HKG Completion Results}\label{sec:HKG completion res}
	We present the benchmark comparison of HKG completion in Table~\ref{tab:main_results}. 
	According to the results, our proposed TransEQ model achieves the state-of-the-art performance on all benchmarks. 
	On the hardest dataset WikiPeople with the most entities and relations, TransEQ significantly improves the best baseline (BoxE) by 27\% and 15\% on Hit@1 and MRR, respectively. 
	Considering hyper-relational connections provided in WikiPeople, this improvement demonstrates that our proposed equivalent transformation preserves complete HKG information. 
	Besides, TransEQ significantly outperforms m-DistMult, its original decoder model without GNN-based encoder, which indicates the effectiveness and necessity to consider structural information in HKG modeling. 
	Such results also imply that with powerful SFs like BoxE, TransEQ can obtain even better performance. 
	Moreover, compared with structural modeling approaches of G-MPNN and StarE, the substantial improvement of TransEQ owes to subtle design of the equivalent transformation as well as the semantic information captured in the decoder part.

\begin{figure}[htbp]
		\centering
		\subfigure[WikiPeople]{ \label{fig:time_wikipeople} 
			\includegraphics[width=0.31\textwidth]{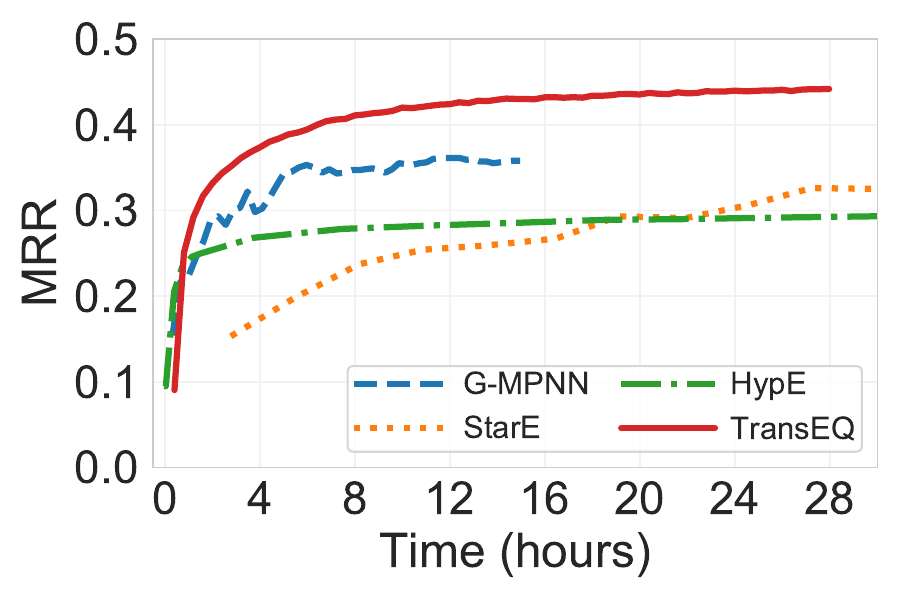}}
		\subfigure[JF17K]{ \label{fig:time_jf17k} 
			\includegraphics[width=0.31\textwidth]{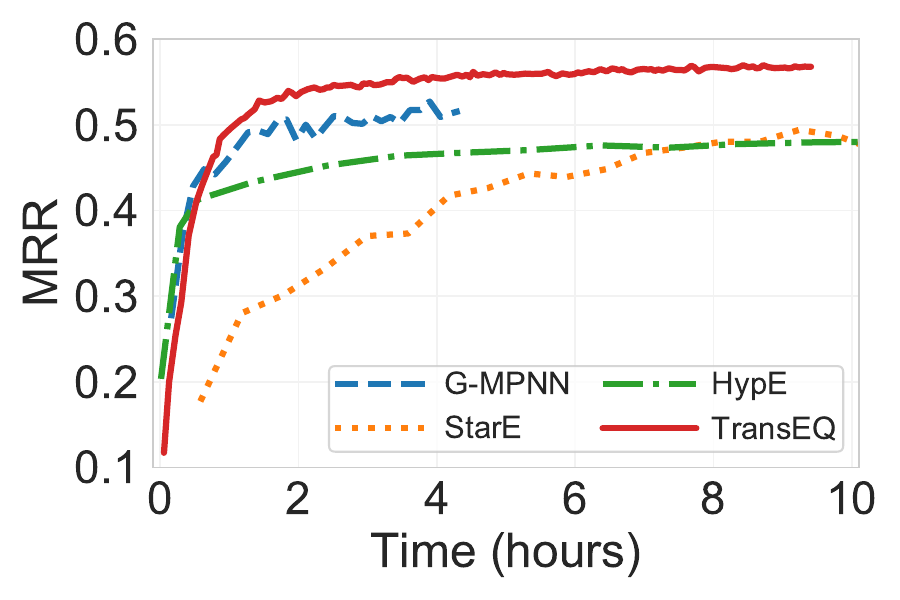}}
		\subfigure[FB-AUTO]{ \label{fig:time_fbauto} 
			\includegraphics[width=0.31\textwidth]{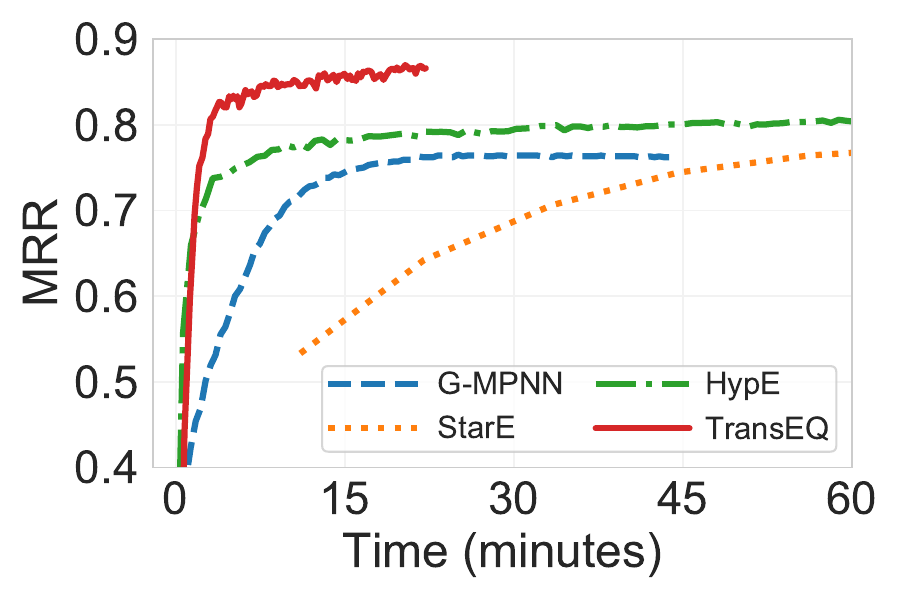}}
		\vspace{-5px}
		\caption{Comparison on clock time of model training vs. testing MRR with structural modeling approaches and HypE.}
		\label{fig:time}
				\vspace{-10px}
	\end{figure}

	Furthermore, we compare the learning processes of TransEQ with structural modeling approaches on three datasets in Figure~\ref{fig:time}. 
	The learning curve of HypE with linear time complexity is also plotted for comparison. 
	It can be observed that TransEQ achieves similar convergence speed with HypE in practice, which owes to the multilinear product based SF \citep{liu2021role} and efficient implementation. 
	With a similar form of SF adopted, G-MPNN achieves a close convergence rate but inferior performance, which demonstrates the strength of GNN-based encoder compared with HGNN. 
	As for StarE with Transformer-based SF, tremendous parameters lead to time-consuming training on all datasets.

	\subsection{Transformation Comparison} 
	To analyze the effects of various transformations, we present the performance comparison in Table~\ref{tab:transformation_results}. 
	As described in Section~\ref{sec:transformation}, our proposed equivalent transformation connects subject and object entities via a relational edge $r$ to form the motif for semantic difference. 
	Thus, we investigate the effectiveness of such operation by removing the edge in the transformation, referred to as w/o distinction transformation.
	
	\begin{table}[htbp]
        \centering
		\caption{Performance comparison of different transformations. Best results are highlighted in bold.}
		\label{tab:transformation_results}
		\def\arraystretch{0.9}
		\begin{tabular}{c|ccc|ccc}
			\toprule
			\multicolumn{1}{c}{} & \multicolumn{3}{c}{\textbf{JF17K}}    & \multicolumn{3}{c}{\textbf{FB-AUTO}}  \\
			\multicolumn{1}{c|}{\textbf{Transformation}} & MRR & Hit@1 & Hit@10 & MRR & Hit@1 & Hit@10 \\
			\midrule
			plain &  0.504   &   0.422 & 0.669 & 0.836 &   0.809  &   0.884     \\
			clique-based plain & 0.557  &  0.478   &  0.706 & 0.824 &  0.804  &   0.859  \\
			clique-based semantic & 0.552 &  0.474 &  0.706 & 0.831 & 0.809 &  0.869 \\
			\midrule
			w/o distinction & 0.536  & 0.456 &  0.695 & 0.849 & 0.810 & 0.884 \\
			equivalent & \textbf{0.569} & \textbf{0.489} & \textbf{0.722} & \textbf{0.870} &   \textbf{0.842} & \textbf{0.909} \\
			\bottomrule
		\end{tabular}
	\end{table}
	
	From the table, we can observe that the equivalent transformation outperforms other variants of transformations, which is in accord with the information loss analysis in Section~\ref{sec:info_loss}, i.e., only equivalent transformation preserves complete information. 
	Moreover, removing the relational edge in the equivalent transformation leads to a Hit@1 performance drop of 7\% on JF17K, which demonstrates the effectiveness and necessity of considering semantic difference in the transformation. 
	Besides, since any two entities are connected in two clique-based transformations, the relatedness between entities is largely captured and thus they obtain close performance, i.e., the clique structure makes these transformations insensitive to semantic information. 
	In comparison with equivalent transformation, the star structure in plain transformation is quite simple without hyper-relational semantics incorporated, which also accounts for the obvious gap between these two transformations. Such results also validate the effectiveness of our generalization from hyperedge expansion to equivalent transformation. Considering the zero information loss and experimental performance, the equivalent transformation becomes the best choice for TransEQ. 
	
	\subsection{Encoder-Decoder Choice Comparison}\label{sec:encoder_decoder}
	To further investigate the effects of different GNN-based encoders along with HKG-based SFs as decoders, we compare the performance of different encoder-decoder choices on JF17K and FB-AUTO in Table~\ref{tab:encoder-decoder}. 
	In the table, each result corresponds to the TransEQ model with \textbf{X} as encoder and \textbf{Y} as decoder.
	
	\begin{table}[htbp]
        \centering
		\caption{Performance comparison of different encoder-decoder choices on JF17K and FB-AUTO. "OOM" indicates out of memory. Best results for each encoder are highlighted in bold.}
		\label{tab:encoder-decoder}
		\def\arraystretch{0.9}
		\begin{tabular}{ccc|cc|cc}
			\toprule
			&	\multicolumn{6}{c}{\textbf{JF17K}}       \\ \cmidrule{2-7} 
			\textbf{Encoder X }$\bm{\rightarrow}$ & \multicolumn{2}{c}{CompGCN} & \multicolumn{2}{c}{R-GCN} & \multicolumn{2}{c}{No Encoder} \\
			\textbf{Decoder Y }$\bm{\downarrow}$ & MRR & Hit@10 & MRR & Hit@10  & MRR & Hit@10     \\ 
			\midrule
			\multicolumn{1}{c|}{m-TransH}  & 0.542 & 0.694 & \multicolumn{2}{c|}{OOM}  & 0.444 & 0.581\\
			\multicolumn{1}{c|}{Transformer} & 0.526   & 0.677 & \multicolumn{2}{c|}{OOM}  & 0.504 & 0.648\\
			\multicolumn{1}{c|}{m-DistMult} & 0.569 & \textbf{0.722}  & 0.483  & 0.632 & 0.452 & 0.599 \\ 
			\multicolumn{1}{c|}{HypE} & \textbf{0.572} & 0.715  & \textbf{0.550} & \textbf{0.702} & \textbf{0.507} & \textbf{0.669} \\
			\bottomrule
		\end{tabular}

		\begin{tabular}{ccc|cc|cc}
			\toprule
			& \multicolumn{6}{c}{\textbf{FB-AUTO}}       \\ \cmidrule{2-7} 
			\textbf{Encoder X }$\bm{\rightarrow}$ & \multicolumn{2}{c}{CompGCN} & \multicolumn{2}{c}{R-GCN} & \multicolumn{2}{c}{No Encoder} \\
			\textbf{Decoder Y }$\bm{\downarrow}$     & MRR & Hit@10  & MRR   & Hit@10  & MRR        & Hit@10     \\ 
			\midrule
			\multicolumn{1}{c|}{m-TransH}  & 0.825 & 0.873 & \multicolumn{2}{c|}{OOM}  & 0.728 & 0.728\\
			\multicolumn{1}{c|}{Transformer} & 0.846   & 0.899 & \multicolumn{2}{c|}{OOM}  &\textbf{0.834} & \textbf{0.897}    \\
			\multicolumn{1}{c|}{m-DistMult}    & 0.870   & \textbf{0.909}   & 0.834  & 0.892   & 0.784      & 0.845   \\ 
			\multicolumn{1}{c|}{HypE} & \textbf{0.860}      & 0.902   & \textbf{0.840}    & \textbf{0.892}   & 0.804  & 0.856      \\
			\bottomrule
		\end{tabular}
	\end{table}
 
	According to the results of each row in Table~\ref{tab:encoder-decoder}, compared with original models (\textbf{X}=No Encoder), TransEQ models with various GNN-based encoders bring substantial improvement, which again demonstrates the effectiveness of structural information encoding. 
	Since neural network models with tremendous parameters easily overfit, the performance improvement of GNN-based encoder for Transformer is much lower than that for other models. 
	As for the encoder in each column, the decoder choices of HypE achieve the best performance, 
	mainly attributed to the linear complexity and full expressivity property. 
	Benefited from the proposed encoder-decoder framework, TransEQ can flexibly adapt to various GNNs and SFs for both superior performance and full expressivity.

	\subsection{Information Sharing Study}\label{sec:information_sharing_study}
	
\begin{figure}[htbp]
		\centering
		\subfigure[$\alpha=0.8$]{ \label{fig:vis_0.8} 
			\includegraphics[width=0.45\textwidth]{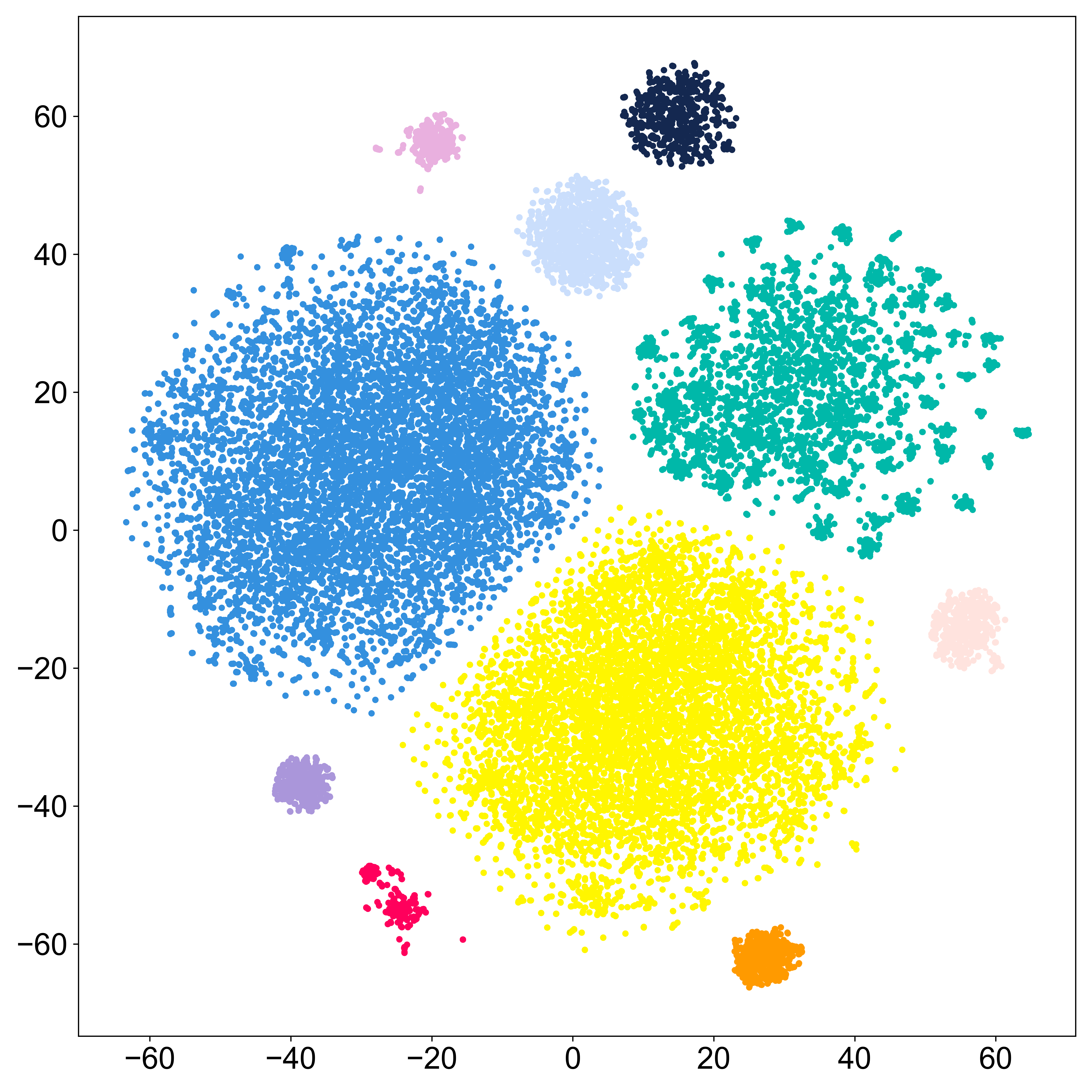}}
		\subfigure[$\alpha=0.0$]{ \label{fig:vis_0.0} 
			\includegraphics[width=0.45\textwidth]{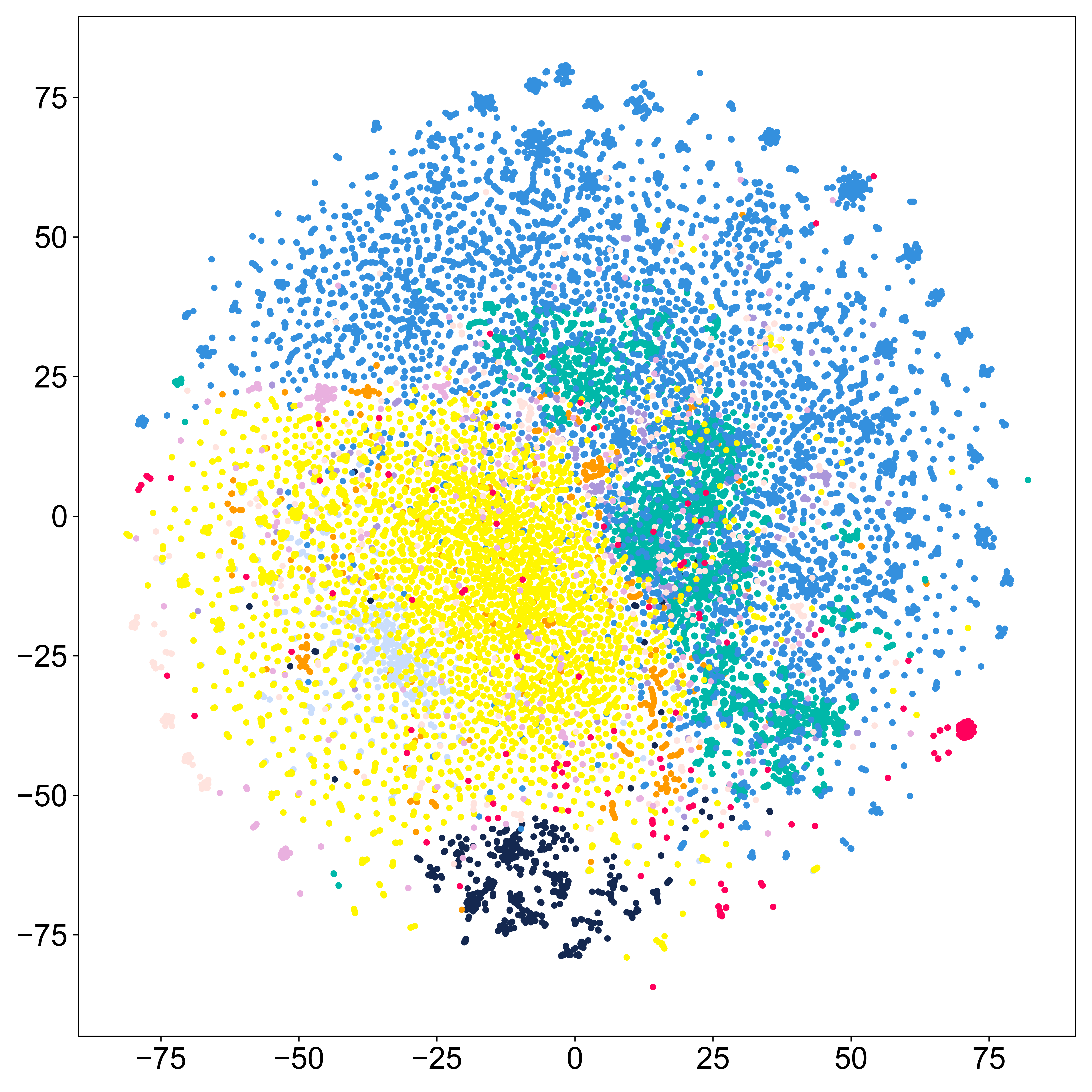}}
				\vspace{-5px}
		\caption{The visualization of mediator entity embeddings with top ten primary relations in WikiPeople via t-SNE on share embedding ratio (a) $\alpha=0.8$ and (b) $\alpha=0.0$.}
		\label{fig:vis_compare}
	\end{figure}

	To validate whether the semantic relatedness in HKG is captured by sharing embedding on mediator entities, we obtain hyper-relational facts of top ten primary relations and visualize their mediator entity embeddings via t-SNE \citep{maaten2008visualizing}, as shown in Figure~\ref{fig:vis_compare}, which compare the cases of utilizing sharing embedding (the best setting with $\alpha=0.8$) and independent embedding ($\alpha=0.0$). We select WikiPeople for visualization considering explicit attribute information therein, and mediator entities belonging to the same primary relation are marked in the same color. 
	From Figure~\ref{fig:vis_compare}(a), we observe that mediator entities are neatly clustered according to their mapping primary relations, which is in accord with our sharing embedding design in GNN-based encoder.  
	In comparison, without the sharing embedding, the learnt mediator entities in Figure~\ref{fig:vis_compare}(b) are hard to be distinguished. 
	
\begin{figure}[htbp]
		\centering
		\subfigure[JF17K]{
			\includegraphics[width=0.4\textwidth]{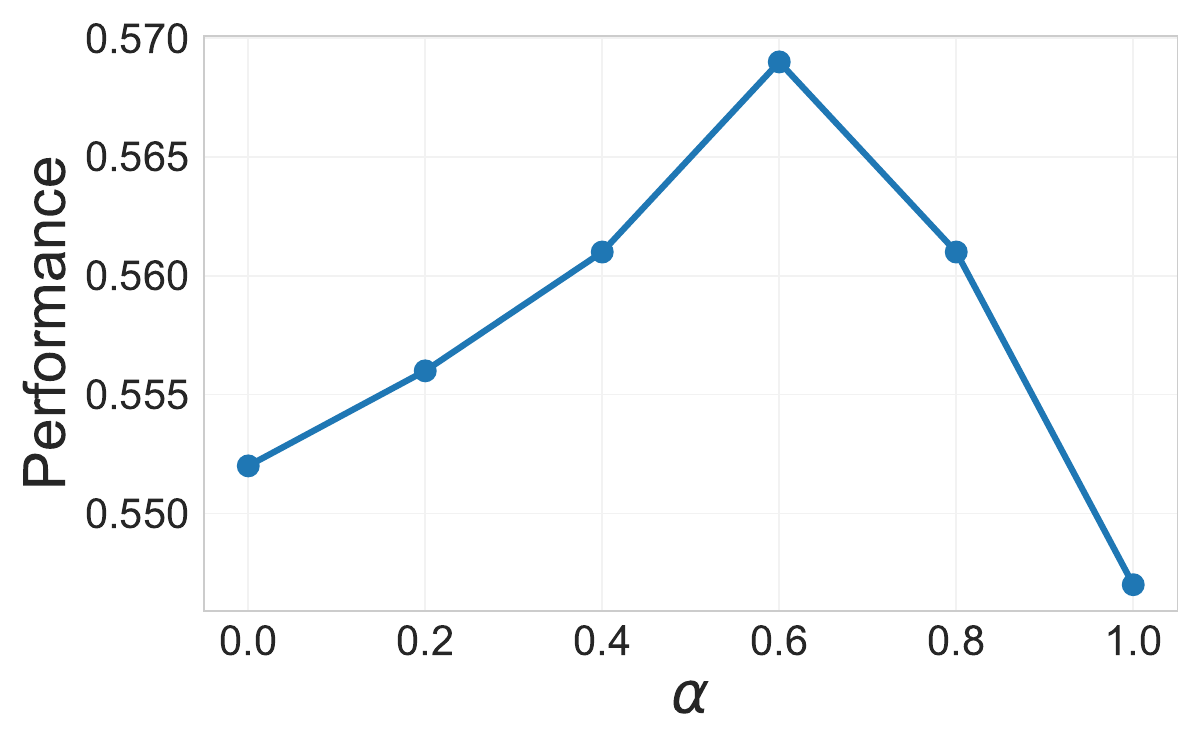} }
		\subfigure[FB-AUTO]{
			\includegraphics[width=0.4\textwidth]{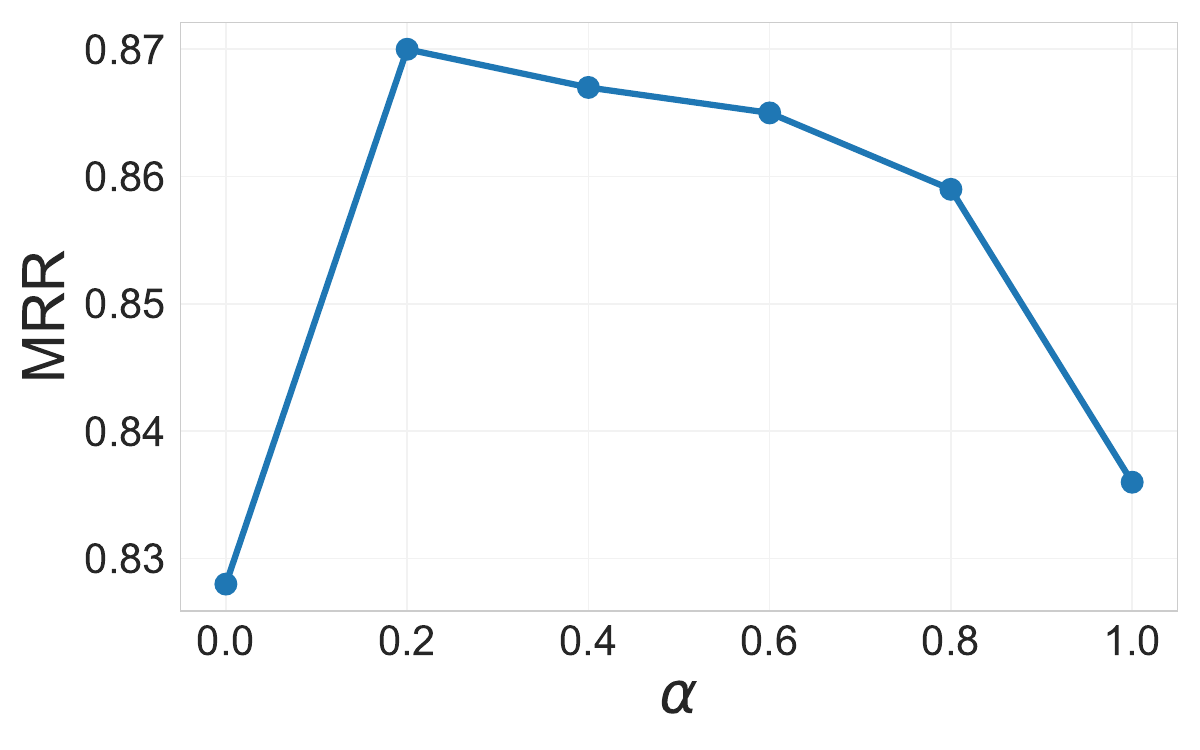}}
		\caption{The effects of share embedding ratio $\alpha$ on (a) JF17K and (b) FB-AUTO.}\label{fig:sharing}
	\end{figure}

	We further investigate the effect of sharing hyperparameter $\alpha$ in Figure~\ref{fig:sharing}(a) and (b). 
	An extreme point can be observed in both datasets, which estimates the semantic relatedness in corresponding datasets. 
	Moreover, a higher sharing ratio $\alpha$ brings fewer model parameters, e.g., $\alpha=0$ corresponds to the case that each mediator entity has independent embedding while $\alpha=1$ means all mediator entities with the same primary relation own the same representation. 
	Thus, a tradeoff between model parameter complexity and practical performance can be achieved.

	\section{Conclusion} \label{sec:conclusion}
	In this paper, we propose TransEQ for HKG modeling.
	By generalizing the hyperedge expansion to the equivalent transformation, TransEQ successfully transforms a HKG to a KG without information loss. 
	Especially, TransEQ builds an encoder-decoder framework, which firstly captures both structural information and semantic information for HKG. 
	Experiment results show that TransEQ obtains the state-of-the-art results on benchmark datasets. 
	
	As future work, the effects of other advanced multi-relational GNNs \citep{yu2021knowledge} and HKG modeling approaches \citep{abboud2020boxe,liu2021role} will be investigated.  
	Besides, we also plan to investigate our proposed TransEQ with inductive setting \citep{ali2021improving} as well as logical query \citep{alivanistos2021query,chen2022explainable}.

\acks{This research has been supported in part by the National Key Research and Development Program of China under Grant 2022ZD0116402;
in part by the National Natural Science Foundation of China under
Grant U22B2057, Grant U20B2060, and Grant 61971267.}

\bibliography{ref}

\newpage

\appendix
	\section{Theoretical Details} \label{app:theoretical details}
	
	\subsection{Complexity Analysis on Transformations}\label{app:transformation_parameter}
	Based on the description in Section~\ref{sec:other_transformation}, we analyze the parameter complexity of different transformations in Table~\ref{tab:complexity}.
	
	\begin{table}[htbp]
        \centering
		\caption{The parameter complexity of different transformations, in terms of entity/node, relation and edge. $n_e=\left|\mathcal{E}\right|$ and $n_r=\left|\mathcal{R}\right|$ are the number of entities and relations in HKG. $n^{\textnormal{pri}}_r$ and $n^{\textnormal{qua}}_r$ are the numbers of primary and qualifier relations, respectively. $n_a$ is the maximum number of attribute-value qualifiers for facts. $N^{\textnormal{qua}}$ and $N^{\textnormal{pri}}$ are the number of hyper-relational facts with and without attribut-value qualifiers, such that $N^{\textnormal{pri}}+N^{\textnormal{qua}}=\left|\mathcal{F}\right|$.}
		\label{tab:complexity}
		\def\arraystretch{0.9}
		\small{
			\begin{tabular}{c|ccc}
				\toprule
				\multicolumn{1}{c|}{\small\textbf{Transformation}} & $\bm{\mathcal{O}_{\textnormal{ent}}/\mathcal{O}_{\textnormal{node}}}$ & $\bm{\mathcal{O}_{\textnormal{edge}}}$ & $\bm{\mathcal{O}_{\textnormal{rel}}}$ \\
				\midrule
				\small plain &  $\mathcal{O}(n_e\!+\!N^{\textnormal{qua}})$   & $\mathcal{O}(N^{\textnormal{pri}}\!+\!N^{\textnormal{qua}}(n_a\!+\!2)\!)$  &  $\mathcal{O}(n_r^{\textnormal{pri}})$  \\
				\midrule
				\small{clique-based plain}   & $\mathcal{O}(n_e)$ & $\mathcal{O}(N^{\textnormal{pri}}\!+\! N^{\textnormal{qua}}(n_a+2)^2)$   &   $\mathcal{O}(n^{\textnormal{pri}}_r)$     \\
				\midrule
				\specialcell{\small clique-based\\ \small semantic} & $\mathcal{O}(n_e)$    & $\mathcal{O}(N^{\textnormal{pri}}\!+\!N^{\textnormal{qua}} (n_a+2)^2)$  &   $\mathcal{O}(n^{\textnormal{pri}}_r\!+\!n_a n^{\textnormal{qua}}_r)$   \\
				\midrule
				\small equivalent &   $\mathcal{O}(n_e\!+\!N^{\textnormal{qua}})$   &    $\mathcal{O}(N^{\textnormal{pri}}\!+\!N^{\textnormal{qua}}(n_a\!+\!3))$   & $\mathcal{O}(3n^{\textnormal{pri}}_r\!+\!n^{\textnormal{qua}}_r)$    \\
				\bottomrule
			\end{tabular}
		}
	\end{table}

	According to the transformation design, clique-based transformations introduce pairwise edges for relatedness while star-based ones of plain transformation and equivalent transformation rely on additional mediator entities. 
	Therefore, clique-based transformations keep the node complexity of $\mathcal{O}(n_e)$ while star-based ones build $\mathcal{O}(n_e\!+\!N^{\textnormal{qua}})$ nodes. 
	On the other hand, the plain transformation keeps the same edge complexity with the original HKG structure, while a relational edge between subject and object entities is added in equivalent transformation for semantic difference, bringing the complexity increase of $\mathcal{O}(N^{\textnormal{qua}})$. 
	Compared with the relation complexity of about $\mathcal{O}(n^{\textnormal{pri}}_r\!+\!n^{\textnormal{qua}}_r)$ in the original HKG, 
	the equivalent transformation introduces $\mathcal{O}(n^{\textnormal{pri}}_r)$ relations to distinguish links between subject and object entities, which are acceptable in practice.

	\subsection{Proof of Information Preservation}\label{app:info_loss}
	To demonstrate the zero information loss in the equivalent transformation, in Algorithm~\ref{alg:star_recover}, we present the process that can equivalently recover original HKG from transformed KG. 
	
	Note that $\mathcal{N}_{b_k}$ in line 3 is a subgraph, and attribute-value qualifiers can be extracted from direct relational links to mediator $b_k$ in line 5. 
	Here we consider hyper-relational fact with at least one qualifier, while triple facts can be directly added in recovered HKG due to no mediator. 
	Thus, Algorithm~\ref{alg:equiv_transformation} and Algorithm~\ref{alg:star_recover} form an equivalent conversion between HKG and KG, i.e., the equivalent transformation preserves the complete information.
	In comparison, the plain transformation and clique-based plain transformation only keep primary relations in conversion, which cannot be recovered due to attribute loss. 
	Besides, clique-based semantic transformation inherits the structural information loss of clique expansion in hyperedge expansion \cite{dong2020hnhn}. 
	
	\SetKwInput{KwInit}{Init}
	\begin{algorithm}[h]
		\caption{The algorithm for recovering HKG from the transformed KG by equivalent transformation.}\label{alg:star_recover}
		
		\KwIn{Transformed KG $\mathcal{G}=(\mathcal{E},\mathcal{R},\mathcal{F})$; }
		
		\KwInit{Recovered HKG $\mathcal{G}^H=(\mathcal{E}^H,\mathcal{R}^H,\mathcal{F}^H)$ with $\mathcal{E}^H\leftarrow\emptyset,$ $\mathcal{R}^H\leftarrow\emptyset,\mathcal{F}^H\leftarrow\emptyset$;}
		
		Obtain the set of mediator entities from $\mathcal{E}$, $\mathcal{E}^{\textnormal{med}}$;\\
		
		\For{$b_k\in\mathcal{E}^{\textnormal{med}}$}
		{
			\raggedright{
				Find $b_k$'s neighbor entities and their connected relations from $\mathcal{F}$, $\mathcal{N}_{b_k}=\{r_i, e_i\}^n_{i=1}$;
			}
			\\
			Extract $(s,r,o)$ from $\mathcal{N}_{b_k}$ via motif-structure discovery;\\
			Extract $\{(a_i,v_i)\}^{n-2}_{i=1}$ from left parts of $\mathcal{N}_{b_k}$;\\
			\tcp{part of $\{r_i,e_i\}^n_{i=1}$ corresponds to $\{(a_i,v_i)\}^{n-2}_{i=1}$.} 
			$\mathcal{E}^H\leftarrow\mathcal{E}^H\cup\{e_i\}^n_{i=1}$, $\mathcal{R}\leftarrow\mathcal{R}\cup\{r\}\cup\{a_i\}^{n-2}_{i=1}$;\\
			$\mathcal{F}^H\leftarrow\mathcal{F}^H\cup\{(s,r,o,\{(a_i,v_i)\}^{n-2}_{i=1})\}$;\\
		}
		\KwOut{Recovered HKG $\mathcal{G}^H=(\mathcal{E}^H,\mathcal{R}^H,\mathcal{F}^H)$.}
	\end{algorithm}
	
	\subsection{Proof of Full Expressivity}\label{app:expressive}
	Our proposed TransEQ firstly transforms a HKG to a KG, then develops a GNN-based encoder for representation encoding, and calculates plausibility scores based on existing SFs in HKG modeling studies with entity and relation embeddings from encoder part. 
	Meanwhile, several SFs from HypE \cite{fatemi2019knowledge}, BoxE \cite{abboud2020boxe}, RAM \cite{liu2021role}, etc., have been proved to be fully expressive with an assignment of entity and relation embeddings in their original papers. 
	Hence, with a fully expressive SF in decoder, the TransEQ model is fully expressive if the output embeddings from encoder part follow corresponding assignment required by SF, which is proved as follows,
	\begin{proof}
		For $t\in\mathcal{E}, r\in\mathcal{R}$, let $\bm{h}^0_{t},\bm{h}^0_{r}$ denote their initialized representations, while $\bm{h}^L_{t},\bm{h}^L_{r}$ denote corresponding embeddings outputted from encoder part. 
		We also denote $\bm{h}^{\textnormal{SF}}_{t},\bm{h}^{\textnormal{SF}}_{r}$ the required input embeddings of SF in decoder. 	
		Then, in mathematical, with $\bm{h}^L_{t}\!=\!Enc(\bm{h}^0_{t},\bm{\theta}_{\textnormal{Enc}})$ and $\bm{h}^L_{r}\!=\!Enc(\bm{h}^0_{t},\bm{\theta}_{\textnormal{Enc}})$, we should prove  $\bm{h}^L_{t}=\bm{h}^{\textnormal{SF}}_{t}$ and $\bm{h}^L_{r}=\bm{h}^{\textnormal{SF}}_{r}$ can be achieved with appropriate choice of encoder parameters $\bm{\theta}_{\textnormal{Enc}}$\footnote{Note that here we simplify the expression of encoder module, which is still in accord with the form in Algorithm~\ref{alg:training}.} and initialized embeddings $\bm{h}^0_{t},\bm{h}^0_{r}$.
		
		Taking the example of R-GCN \cite{schlichtkrull2018modeling} as encoder, the message passing process of each GCN layer can be written as,
		\begin{align}
			\bm{h}^{l+1}_t=\sigma(\sum_{r\in\mathcal{R}}\sum_{u\in\mathcal{N}^r_t}\frac{1}{\left|\mathcal{N}^r_t\right|}\bm{W}^{l}_r\bm{h}^{l}_u+\bm{W}^l_0\bm{h}^l_t),\notag
		\end{align}
		where $\sigma$ denotes nonlinear activation function like ReLU, which is unnecessary and can be removed \cite{wu2019simplifying}. 
		
		Now, we describe a feasible assignment of encoder parameters: For each layer $l\in\{1,\cdots,L\}$ and $r\in\mathcal{R}$, relation-specific matrix $\bm{W}^l_r$ is set to null matrix, while $\bm{W}^l_0$ is set to identity matrix, where both $\bm{W}^l_r$ and $\bm{W}^l_0$ belong to encoder parameters $\bm{\theta}_{\textnormal{Enc}}$.
		
		Following the assignment above, we have $\bm{h}^{l+1}_t=\bm{h}^l_{t}$, i.e., $\bm{h}^L_{t}=\bm{h}^0_{t}$. 
		Hence, we can set the values of $\bm{h}^0_t$ according to $\bm{h}^{SF}_t$. 
		In R-GCN, $\bm{h}^L_{r}$ is directly initialized and can be set to $\bm{h}^{\textnormal{SF}}_r$. 
		Overall, the encoder's output embeddings follow the required embedding assignment of SF with above assignment on  $\bm{\theta}_{\textnormal{Enc}}$, $\bm{h}^0_{t}$ and $\bm{h}^0_{r}$. 
		Thus, the expressivity of TransEQ is proved to be in accord with that of the SF it uses in decoder. 
		Finally, we note that the proof can be trivially extended to other GNN-based encoders like CompGCN \cite{vashishth2019composition} by introducing extra assignments on encoder parameters.
	\end{proof}
	
	\section{Dataset Processing}\label{app:exp_details}
	
	As described in the introduction, each hyper-relational fact contains a primary triple and several attribute-value qualifiers. However, some experiment datasets like JF17K and FB-AUTO did not provide the information of primary and qualifier parts. Therefore, following the dataset processing in \citep{StarE,rossobeyond,guan2020neuinfer}, we identify such two parts for samples in corresponding datasets.
	
	For JF17K and FB-AUTO datasets, the dataset processing is as follows. For better understanding, here we simplify the expression with some abuse of notation.
	\begin{itemize}[leftmargin=10px]
		\item Raw data format: $(r_{raw},e_s,e_o,e_1,\cdots,e_n)$.
		\item Identification:The first two entities in the raw data are default subject and object entities. We split the (n+2)-ary relation $r_{raw}$
		into $r^{so}, r_1,\cdots,r_n$
		.
		\item Converted data format: $(e_s,r^{so},e_o,\{(r_i:e_i)\}^n_{i=1})$ 
	\end{itemize}
	
	An example in JF17K dataset processing is as follows.
	\begin{itemize}[leftmargin=10px]
		\item Raw data sample: (soccer.football\_player\_match\_participation, 02pp1, 04xkpd, 0c0lv1g)
		\item Identification:The primary triple (02pp1, soccer.football\_player\_match\_ participation\_so, 04xkpd)
		\item Converted data sample: (02pp1, soccer.football\_player\_match\_partici pation\_so, 04xkpd, (soccer.football\_player\_match\_participation\_1, 0c0lv1g))
	\end{itemize}
	
	The WikiPeople dataset provides the information of primary and qualifier parts, and the dataset processing is as follows.
		\begin{itemize}[leftmargin=10px]
		\item Raw data format: $\{r^{sub}: e_s,r^{obj}:e_o,r_1:e_1,\cdots,r_n:e_n\}$.
		\item Identification:Since the raw data provide the subject and object entities, we combine $r^{sub}$
		and $r^{obj}$
		to a new primary relation $r$. 
		\item Converted data format: $(e_s,r,e_o,\{(r_i:e_i)\}^n_{i=1})$ 
	\end{itemize}
	
	An example in WikiPeople dataset processing is as follows.
		\begin{itemize}[leftmargin=10px]
		\item Raw data sample: \{P3919\_h:Q337913, P3919\_t:Q1210343, P2868: Q864380\}
		\item Identification:The primary triple (Q337913, P3919, Q1210343)
		\item Converted data sample: (Q337913, P3919, Q1210343, (P2868,Q864380))
	\end{itemize}
	Following the dataset processing above, we guarantee the fair comparison of baselines and our proposed model.

\end{document}